%% file: main.tex
\documentclass[lettersize,journal]{IEEEtran}
\usepackage{amsmath,amsfonts}
\usepackage{algorithm}
\usepackage{array}
\usepackage[caption=false,font=normalsize,labelfont=sf,textfont=sf]{subfig}
\usepackage{textcomp}
\usepackage{stfloats}
\usepackage{url}
\usepackage{verbatim}
\usepackage{graphicx}
\usepackage{hyperref}
\usepackage{multirow}
\usepackage{algpseudocode}
\usepackage{soul}
\usepackage{color, xcolor}
\usepackage{cite}
\usepackage[numbers]{natbib}
\begin{document}

\title{CycleDiff: Cycle Diffusion Models for Unpaired Image-to-image Translation}

\author{Shilong Zou$^{\dagger}$, Yuhang Huang$^{\dagger}$\thanks{$\dagger$ Equal contribution.}, Renjiao Yi, Chenyang Zhu$^{*}$, 

Shixiang Wang, Xiangchao Zhang, Kai Xu$^{*}$\thanks{$*$ Co-corresponding author 

},~\IEEEmembership{Senior Member,~IEEE}
\thanks{S. Zou, Y. Huang, R. Yi and C. Zhu are with the College of Computer Science and Technology, National University of Defense Technology, Changsha 410073, China (e-mail: zoushilong@nudt.edu.cn;huangai@nudt.edu.cn; yirenjiao@nudt.edu.cn;zhuchenyang07@nudt.edu.cn). Xiangchao Zhang and Shixiang Wang are with College of Future Information Technology, Fudan University, Shanghai 200438, China.

K. Xu is with the School of Computer, National University of Defense Technology, Changsha 410073, China, also with the Xiangjiang Laboratory, Changsha 410013, China, and also with the Institute of AI for Industries, Chinese Academy of Sciences, Nanjing 211135, China
(e-mail: kevin.kai.xu@gmail.com)}

\href{https://zoushilong1024.github.io/CycleDiff/}{CycleDiff.github.io}
}

\markboth{Journal of \LaTeX\ Class Files,~Vol.~14, No.~8, August~2021}%
{Shell \MakeLowercase{\textit{et al.}}: A Sample Article Using IEEEtran.cls for IEEE Journals}


\maketitle

\begin{abstract}
We introduce a diffusion-based cross-domain image translator in the absence of paired training data.
Unlike GAN-based methods, our approach integrates diffusion models to learn the image translation process, allowing for more coverable modeling of the data distribution and performance improvement of the cross-domain translation.
However, incorporating the translation process within the diffusion process is still challenging since the two processes are not aligned exactly, i.e., the diffusion process is applied to the noisy signal while the translation process is conducted on the clean signal. 
As a result, recent diffusion-based studies employ separate training or shallow integration to learn the two processes, yet this may cause the local minimal of the translation optimization, constraining the effectiveness of diffusion models.
To address the problem, we propose a novel joint learning framework that aligns the diffusion and the translation process, thereby improving the global optimality.
Specifically, we propose to extract the image components with diffusion models to represent the clean signal and employ the translation process with the image components, enabling an end-to-end joint learning manner.
On the other hand, we introduce a time-dependent translation network to learn the complex translation mapping, resulting in effective translation learning and significant performance improvement.
Benefiting from the design of joint learning, our method enables global optimization of both processes, enhancing the optimality and achieving improved fidelity and structural consistency.
We have conducted extensive experiments on RGB$\leftrightarrow$RGB and diverse cross-modality translation tasks including RGB$\leftrightarrow$Edge, RGB$\leftrightarrow$Semantics and RGB$\leftrightarrow$Depth, showcasing better generative performances than the state of the arts.
{Especially, our method achieves the best FID score in widely-adopted tasks and outperforms the second-best method with an improved FID of 19.61 and 19.67 on Dog$\to$Cat and Dog$\to$Wild respectively.}
\end{abstract}


\input{sections/intro}
\input{sections/related}
\input{sections/method}
\input{sections/exp}
\input{sections/conclu}
\input{sections/ack}

\bibliographystyle{IEEEtran}
\bibliography{IEEEabrv,ref}


\vspace{11pt}

\begin{IEEEbiography}[{\includegraphics[width=1in,height=1.25in, clip,keepaspectratio]{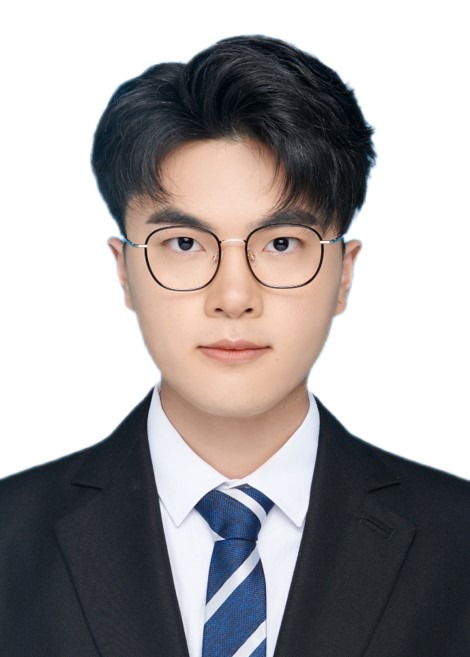}}]{Shilong Zou} received the bachelor's degree from the University of Dalian Maritime University, Dalian, 2023. He is currently pursuing the master's degree at the National University of Defense Technology, Changsha, China. 

His research interests include computer vision and generative models.
\end{IEEEbiography} 

\vspace{11pt}

\begin{IEEEbiography}[{\includegraphics[width=1in,height=1.25in,clip,keepaspectratio]{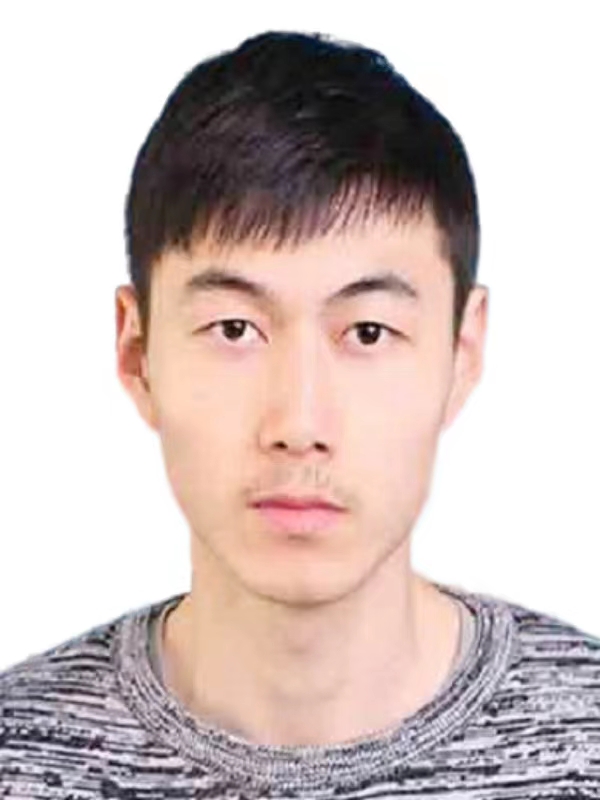}}]{Yuhang Huang}
received the bachelor’s degree from the University of Shanghai for Science and Technology, Shanghai, China, in 2019, and the master’s degree from Shanghai University, Shanghai, China, in 2022. He is currently pursuing the Ph.D. degree at the National University of Defense Technology, Changsha, China.

His research interests include computer vision, graphics, and generative models.
\end{IEEEbiography}

\vspace{11pt}
\begin{IEEEbiography}[{\includegraphics[width=1in,height=1.25in,clip,keepaspectratio]{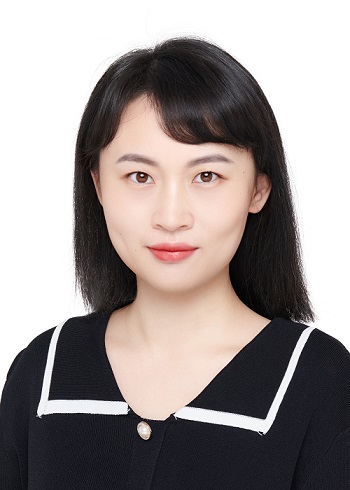}}]{Renjiao Yi}
is an Associate Professor at the School of Computer, National University of
Defense Technology. She received her Ph.D. degree from Simon Fraser University in 2019. She is interested in inverse rendering, 3D scene understanding \& editing, and related AR applications.
\end{IEEEbiography}

\vspace{11pt}
\begin{IEEEbiography}[{\includegraphics[width=1in,height=1.25in,clip,keepaspectratio]{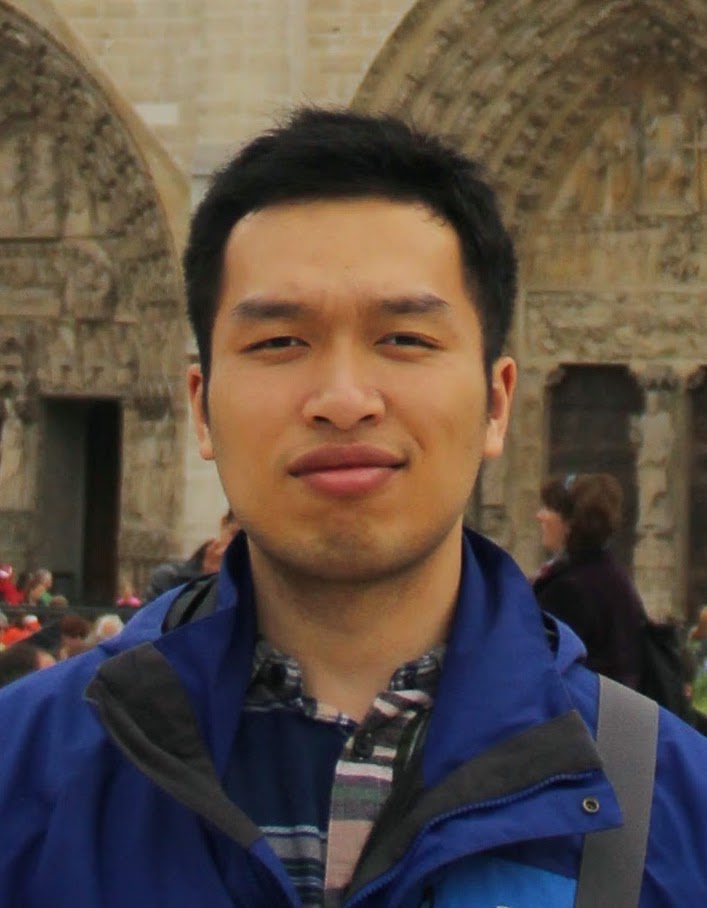}}]{Chenyang Zhu}
is an Associate Professor at the School of Computer, National University of Defense Technology. The current directions of interest include data-driven shape analysis and modeling, 3D vision and, robot perception \& navigation, etc.
\end{IEEEbiography}

\vspace{11pt}
\begin{IEEEbiography}[{\includegraphics[width=1in,height=1.25in,clip,keepaspectratio]{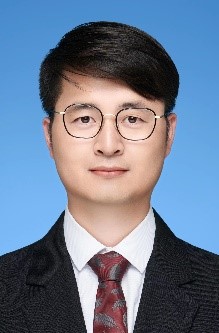}}]{Shixiang Wang}
received the B.Eng. degree from Shanghai Jiao Tong University, China, in 2013 and M.Phil. degree from University of Chinese Academy of Sciences, China, and Ph. D degree from The Hong Kong Polytechnic University, Hong Kong, in 2019. After he received the Ph. D degree, he did the Postdoctoral work at the State Key Laboratory of Ultra-precision Machining Technology, Department of Industrial and Systems Engineering, The Hong Kong Polytechnic University. He is currently an assistant professor in Fudan University. His current research interests include advanced manufacturing technology, precision surface measurement and freeform characterization, etc.
\end{IEEEbiography}

\vspace{11pt}
\begin{IEEEbiography}[{\includegraphics[width=1in,height=1.25in,clip,keepaspectratio]{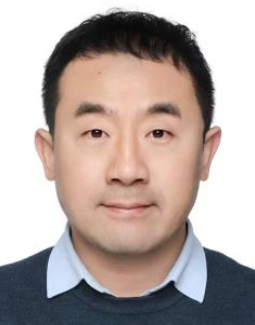}}]{Xiangchao Zhang}
received the B.E degree in measurement technology from the University of Science and Technology of China in 2005 and the Ph. D. degree in precision measurement and instrumentation from the University of  Huddersfield, UK in 2009. Since 2011, he was with the Department of Optical Science and Engineering, Fudan University as an associate professor and has been with the College of Future Information Technology, Fudan University as a full professor since December 2022. He is a senior member of SPIE, and a member of ISO TC213, IEEE and OSA. His research interests include optical measurement technology, micro/nano optics and image processing.
\end{IEEEbiography}

\vspace{11pt}
\begin{IEEEbiography}[{\includegraphics[width=1in,height=1.25in,clip,keepaspectratio]{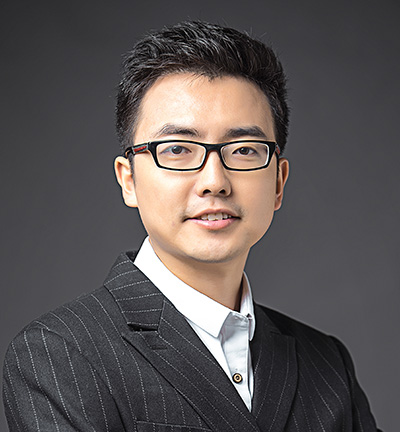}}]{Kai Xu} (Senior Member, IEEE) received the Ph.D. degree in computer science from the National University of Defense Technology (NUDT), Changsha,
China, in 2011. From 2008 to 2010, he worked as a Visiting Ph.D. degree with the GrUVi Laboratory, Simon Fraser University, Burnaby, BC, Canada. He is currently a Professor with the School of Computer Science, NUDT. He is also an Adjunct Professor with Simon Fraser University. His current research interests include data-driven shape analysis and modeling, and 3-D vision and robot perception and navigation.
\end{IEEEbiography}

\vfill

\end{document}

%% file: sections/intro.tex
\section{Introduction}

    \begin{figure}[t]
    \centering
    \includegraphics[width=1\linewidth]{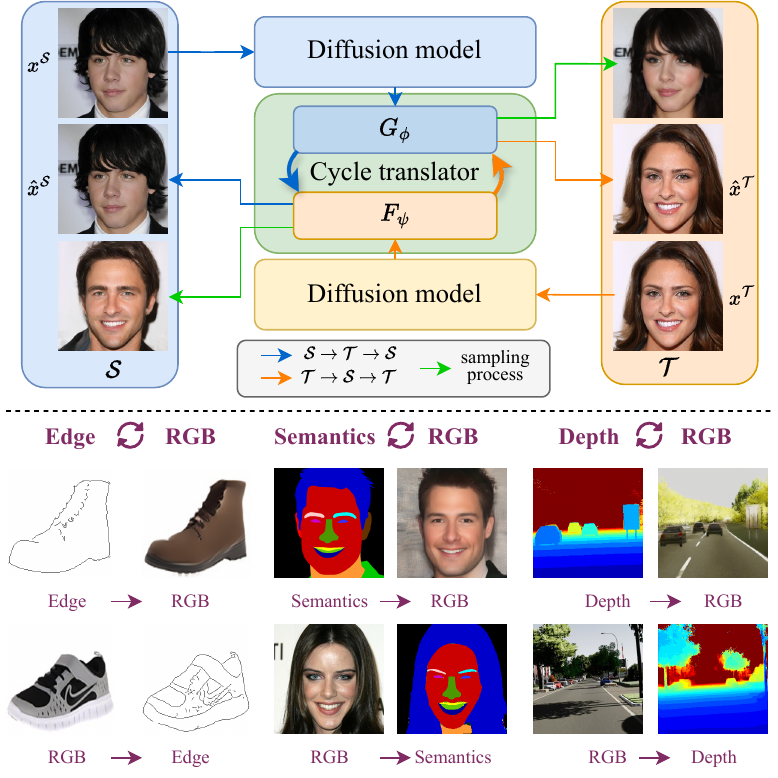}
    \caption{
    The proposed CycleDiff consists of two domain-specific diffusion models and a cycle translator, and learns the diffusion and translation processes jointly.
    The cycle translator consists of two translation network used for performing cycle translation between two domains: $G_\phi:\mathcal{S}\to\mathcal{T}$ and $F_\psi: \mathcal{T}\to\mathcal{S}$. We employ the cycle consistency constrain to regularize the forward and backward translation mappings. 
    Utilizing only unpaired images, CycleDiff can synthesize structure-consistent and photo-realistic results across different modalities of images.}
    \label{fig_teaser}
    \end{figure}

    \IEEEPARstart{I}{mage}-to-image translation is an important and useful task in computer vision and has been attracting increasing attention lately. Generalized image-to-image tasks include image style translation \cite{richardson2021encoding, wang2020multi}, edge detection \cite{ye2024diffusionedge, liu2020adaptive}, semantic segmentation \cite{zhou2021gmnet, wu2023conditional}, and depth estimation \cite{ye2021unsupervised, xu2021multi}. Common to these tasks is the requirement of paired exemplars (e.g. rgb$\leftrightarrow$edge pairs) for training; obtaining such paired data is often effort intensive.
    For example in Fig.~\ref{fig_teaser}, the desired output is not even well-defined for male$\leftrightarrow$female transfiguration and the pixel-level semantic maps are difficult to annotate. 
    Therefore, leveraging unpaired data solely holds great potential in alleviating the burden of data collection and facilitating the generation of a virtually endless stream of new paired data.

    The typical solution of unpaired image-to-image translation is CycleGAN \cite{zhu2017cyclegan}, which utilizes two generators to conduct the two-sided translation and proposes the cycle-consistency loss to maintain structure consistency.
    The unpaired setting of CycleGAN inspires a series of GAN-based studies \cite{yi2017dualgan, kim2017discogan} for unpaired image-to-image translation. 
    However, they can hardly achieve photo-realistic results due to several reasons. 
    Firstly, the GAN-based generator employs a succinct framework yet does not optimize the distribution loss \cite{kurz2016kullback}, resulting in frequent collapse of image translation.
    Moreover, the complex image-to-image translation is regarded as a one-step mapping in GAN-based methods, i.e., they only call the model function once per generation. However, such one-step mapping always overfits the individual points of the data distribution rather than the entire data distribution.
    As a result, the modeling of the cross-domain translation is hard to cover the distributions of both domains, leading to unsatisfactory translation outcomes.
    This serves as motivation for exploring more robust solutions in unpaired image-to-image translation.

    Recent diffusion probabilistic models have demonstrated superior image generation capabilities to GANs, prompting increased interest in diffusion-based methods for unpaired image-to-image translation.
    EGSDE \cite{zhao2022egsde}, CycleDiffusion \cite{wu2023cyclediffusion} and SDDM \cite{sun2023sddm} focus on revising the sampling equation of diffusion models for translating images between different domains. They only consider the optimization of diffusion but without explicit learning of translation, which hardly improves the cross-domain translation performance. 
    UNIT-DDPM \cite{sasaki2021unit} introduces a learnable translation module between different domains to learn the translation process. However, this module is trained independently from the diffusion-based generation in each domain, which may still lead to sub-optimal translation performance without joint optimization.
    Another line of work attempts to train the diffusion and translation processes simultaneously. 
    SynDiff \cite{ozbey2023unsupervised} trains the two processes jointly; however, the joint optimization is only effective for a single denoising step in the diffusion process, indicating suboptimal convergence.
    Consequently, developing a joint learning framework that involves deeper interaction between the two processes becomes essential to enhance both global optimality and generative quality.

    In this paper, we present \textbf{Cycle} \textbf{Diff}usion Models (CycleDiff) that incorporate the cycle-consistency learning process with the diffusion models. 
    Different from the one-step mapping of GAN-based methods, we learn the translation process with a multi-step mapping, i.e., we conduct the translation process in each denoising step of diffusion models. 
    In this way, our model is more suitable for modeling the entire data distribution of both domains, facilitating the cross-domain translation.
    Compared to previous diffusion-based methods, our approach benefits from deeper integration of joint diffusion and cycle translation, improving the global optimality and maintaining the structure consistency to a great extent.
    Moreover, our method can be easily applied to tackle cross-modality image translation such as RGB$\leftrightarrow$Edge and RGB$\leftrightarrow$Semantics.
    
\begin{figure*}[tb]
    \centering
    \includegraphics[width=1\linewidth]{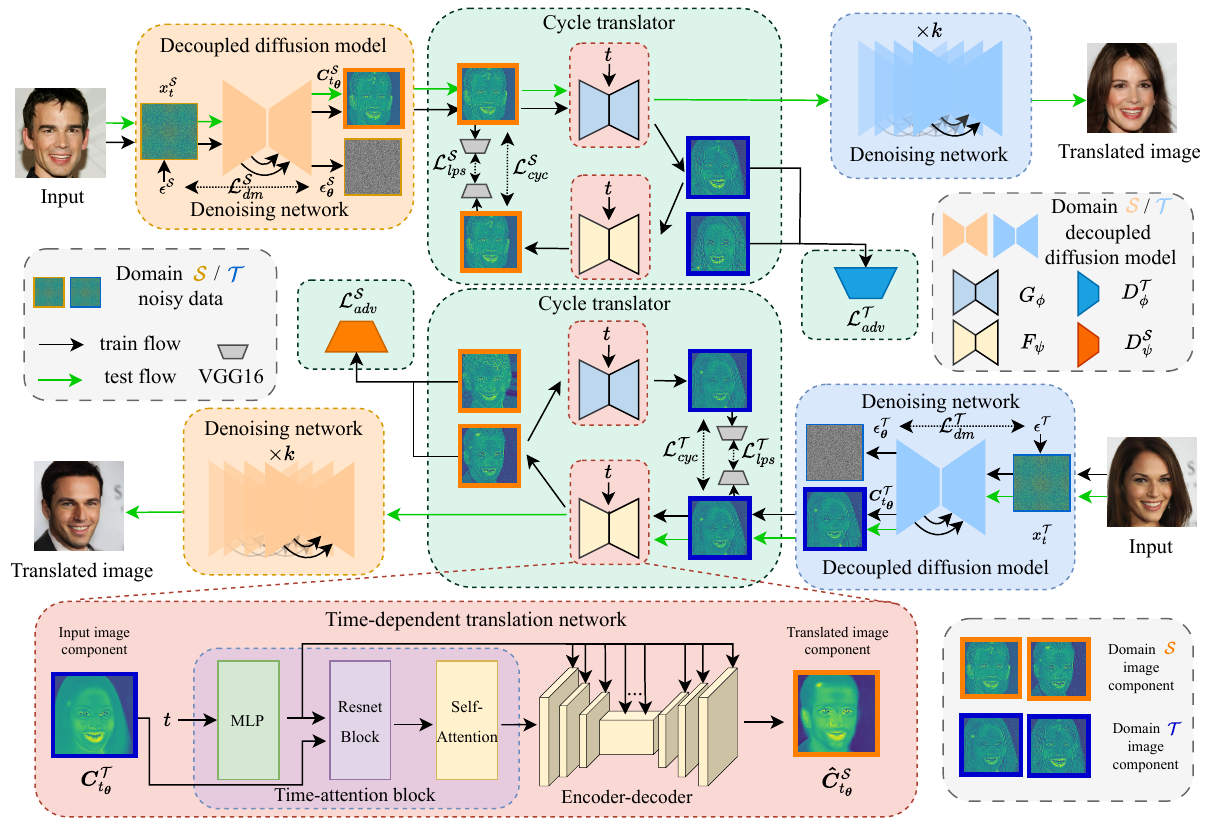}
    \caption{The overall architecture of CycleDiff. CycleDiff comprises two parts: the diffusion models and the cycle translator. The diffusion models are employed to extract image components, which are then fed into the cycle translator for unpaired translation between two domains. The diffusion and translation processes are learned jointly.
    }
    \label{fig_architecture}
    \end{figure*}

    Specifically, as shown in Fig.~\ref{fig_architecture}, we propose to embed a cycle translator between the diffusion models of two different domains, resulting in a joint training framework.
    There are two crucial designs to make the cycle translator work well.
    On the one hand, we extract the image components with the diffusion models and conduct the cycle translation process on the image components. This implementation makes it possible to learn the diffusion and translation processes jointly.
    On the other hand, we introduce a time-dependent translation network that fits the multi-step translation mapping effectively, thus improving translation quality significantly.
    We have conducted extensive experiments on four types of tasks including RGB$\leftrightarrow$RGB, RGB$\leftrightarrow$Edge, RGB$\leftrightarrow$Semantics, and RGB$\leftrightarrow$Depth, achieving new state-of-the-art performances and demonstrating the superiority of the proposed method.

    Our contributions are summarized as follows:
        \begin{itemize}
            \item {We propose a novel joint learning framework for unpaired image-to-image translation that integrates cycle-consistent translation at every denoising step of the diffusion process. This deep integration enhances global optimality while improving generative quality.}
            \item We introduce two key techniques to facilitate the joint learning of translation and diffusion.
            1) we extract the image component from diffusion models to align the input of the translation process, thus allowing for joint learning;
            2) we introduce a time-dependent translation network to improve the translation performance significantly.
            \item {Our method can be easily extended to cross-modality image translation. Extensive experiments show that the proposed method outperforms the state-of-the-art methods on RGB$\leftrightarrow$RGB (including Cat$\leftrightarrow$Dog, Wild$\leftrightarrow$Dog, Male$\leftrightarrow$Female, Old$\leftrightarrow$Young, Summer$\leftrightarrow$Winter, Label$\leftrightarrow$Cityscape, Map$\leftrightarrow$Satellite, Horse$\leftrightarrow$Zebra) and achieves impressive performances on cross-modality translation tasks, including RGB$\leftrightarrow$Edge, RGB$\leftrightarrow$Semantics, and RGB$\leftrightarrow$Depth.}
        \end{itemize}

%% file: sections/related.tex
\section{Related Work}

\textbf{Unpaired image-to-image translation}. Currently, GAN-based methods have achieved impressive results in the realm of paired image-to-image translation tasks \cite{isola2017image}. Nonetheless, there is significant potential for improvement, particularly within the domain of unpaired image-to-image translation, which can be broadly categorized into two-sided and one-sided mapping approaches. For the former framework, the most widely adopted method is the cycle consistency constraint, which limits the translated image to be able to translate back by inverse mapping, including CycleGAN \cite{zhu2017cyclegan}, DualGAN \cite{yi2017dualgan} and DiscoGAN \cite{kim2017discogan}. Following this, there are several studies have been devoted to improving it. U-GA-IT \cite{ugatit} incorporates an attention mechanism to let the models focus on the more important regions distinguishing between domains via the auxiliary classifier. Santa~\cite{xie2023unpaired} introduces the shortest path regularization to find a proper mapping between two different domains based on GAN. However, the GAN-based methods can easily trap in model collapse due to the succinct framework and one-step mapping.

\textbf{Diffusion probabilistic models}. Recently, diffusion probabilistic models have gained popularity as a result of their powerful generative capability through iterative denoising in various fields like images \cite{dhariwal2021diffusionbeatgan, ozdenizci2023restoring}, graph \cite{luo2023fast} and speech \cite{lu2022conditional}. DDPM \cite{ho2020ddpm} first proposed a denoising diffusion probabilistic model, which requires many sampling timesteps to generate high-quality results. Following this, DDIM \cite{nichol2021ddim} aimed to accelerate sampling with implicit probabilistic models. Later, \cite{scorebaseddm} formulated the general SDE framework, which linked the score-based generative model and diffusion probabilistic model. \cite{rombach2022high} incorporate auto-encoder with the diffusion model, allowing the generation of higher-resolution.

\textbf{Diffusion models for unpaired image-to-image translation}. There are lots of work foucsing on the image-to-image translation task. ILVR \cite{ILVR} and SDEdit \cite{SDEdit} leverage score-based diffusion model to guide the generation process based on the target domain images, but both ignore the source domain data. EGSDE \cite{zhao2022egsde} employs an energy function pretrained on the source and target domain to refine the inference process and maintain the faithfulness of translated images. SDDM \cite{sun2023sddm} explicitly optimizes the intermediate generative distributions by decomposing the score function into the ``denoising" part and ``refinement" part, which achieves promising results. CycleDiffusion \cite{wu2023cyclediffusion} defines a latent space for stochastic diffusion model, which enables the unpaired I2I translation, image editing, and plug-and-play guidance with the pretrained diffusion model. All the methods mentioned above lack unpair training between the source and target domain, making it hard to obtain better results for structure similarity. UNIT-DDPM \cite{sasaki2021unit} utilizes two diffusion models and two translation models using cycle consistency constraints to achieve unpaired image-to-image translation, yet without joint learning. UNSB~\cite{UNSB} aims to learn the diffusion-based generation and translation processes jointly via introducing the neural Schrodinger Bridge, however, it focuses on one-sided translation instead of cycle translation.
SynDiff~\cite{ozbey2023unsupervised} also learns the diffusion and cycle translation processes jointly, yet it formulates the translation process as a one-step mapping operation performed exclusively on clean signals.
Unlike the above methods, we learn the translation process with a multi-step mapping, and integrate the translation process into each denoising step of diffusion models, enhancing the global optimality. 

%% file: sections/method.tex
\section{Method}

CycleDiff comprises two main components as depicted in Fig.~\ref{fig_architecture}. In Sec.~\ref{3.1}, we introduce the overall framework of Cycle Diffusion Models, including joint translation in Sec.~\ref{3.1.1} and our proposed Time-dependent translation network in Sec.~\ref{3.1.2}. Then we present the training process and the inference process of CycleDiff in Sec.~\ref{3.2} and Sec.~\ref{3.3} respectively. Finally, we describe the implementation details of the method in Sec.~\ref{3.4}.

\subsection{Problem Formulation}
Our goal is to translate images between two domains $\mathcal{S}$ and $\mathcal{T}$. 
Given two sets of data $\mathcal{S} = \{x_i^\mathcal{S}\}_{i=1}^{N}$ and $\mathcal{T} =\{x_i^\mathcal{T}\}_{i=1}^M$, where $x^\mathcal{S}_i$ and $x^\mathcal{T}_i$ denote the $i$-th images from $\mathcal{S}$ and $\mathcal{T}$ respectively, we translate $x^\mathcal{S}_i$ to $\mathcal{T}$ and $x^\mathcal{T}_i$ to $\mathcal{S}$ simultaneously.

\subsection{Framework of Cycle Diffusion Models}\label{3.1}
\subsubsection{Joint cycle translation and diffusion}\label{3.1.1}
Different from previous diffusion-based methods that only revise the sampling equation or separately train translation and diffusion networks, we propose to learn the two processes jointly, obtaining a unified optimization objective and improving the generative quality greatly.

\begin{figure}[t]
    \centering
    \includegraphics[width=0.49\textwidth]{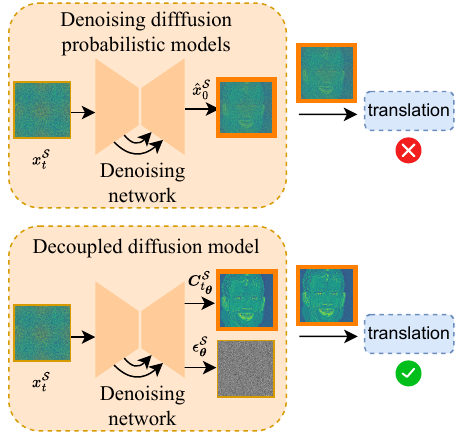}
    \caption{Comparison of decoupled diffusion model and traditional diffusion model \cite{ho2020ddpm} (with `x0 prediction'). The decoupled diffusion model can isolate clean components from the noisy input, while the estimates of the traditional diffusion model are still noisy and not suitable for the subsequent transformation process.}
    \label{fig_adm_other}
    \vspace{-3pt}
\end{figure}

First and foremost, there is a significant issue that must be addressed regarding joint learning. Intuitively, to train the translation process, we need to input the original images from domains $\mathcal{S}$ and $\mathcal{T}$ and learn the mapping between the two domains. 
On the other hand, traditional diffusion models \cite{ho2020ddpm} require estimation of either noise or clean signals from noisy inputs; however, precise estimation remains challenging as noise and clean signals are intricately mixed in the diffusion process. As shown in the top panel of Fig.~\ref{fig_adm_other}, the clean signal estimated by traditional diffusion models \cite{ho2020ddpm} lacks sufficient clarity for subsequent translation. Consequently, directly combining the two training processes without proper alignment between them results in suboptimal performance.

To solve the problem, we propose to extract the image components from the diffusion model to represent the clean images of the two domains. In this way, we can use the image components as the input of the translation process, performing cycle translation and diffusion training jointly. Inspired by decoupled diffusion models \cite{huangddm}, we utilize the denoising network to predict the gradient of the image attenuation at each time $t$, which can be represented as the image component.

Specifically, the forward diffusion process describes the image attenuation and the noise growth processes, which are formulated as:
\begin{equation}
    \boldsymbol{\mathrm{x}}_t^\mathcal{S} = \boldsymbol{\mathrm{x}}_0^\mathcal{S} + \int_0^t \boldsymbol{C}^\mathcal{S}_t \mathrm{d}t + t\boldsymbol{\epsilon}^\mathcal{S},
    \label{eq1}
\end{equation}
Here, $\boldsymbol{\mathrm{x}}_0^\mathcal{S} + \int_0^t \boldsymbol{C}^\mathcal{S}_t \mathrm{d}t$ denotes the image attenuation process and describes the increasing process of noise, and $\boldsymbol{\epsilon}^\mathcal{S}\sim\mathcal{N}(\mathbf{0}, \mathbf{I})$ is the standard normal noise. Essentially, $\boldsymbol{C}^\mathcal{S}_t$ is the gradient of the image attenuation and the image must attenuate to zero when $t=1$, therefore, we can easily obtain $\boldsymbol{C}^\mathcal{S}_t$ via $\boldsymbol{\mathrm{x}}^\mathcal{S}_0+\int_0^1\boldsymbol{C}^\mathcal{S}_t\mathrm{d}t = \mathbf{0}$, i.e., $\boldsymbol{C}^\mathcal{S}_t = -\boldsymbol{\mathrm{x}}^\mathcal{S}_0$. Obviously, the image attenuation process is governed by $\boldsymbol{C}^\mathcal{S}_t$, thus we can use $\boldsymbol{C}^\mathcal{S}_t$ as the image component. In the training process, we use the denoising network to estimate the image component additionally. In practice, the noisy image $\boldsymbol{\mathrm{x}}^\mathcal{S}_t$ is fed into the denoising network together with the time step $t$, outputting the predicted noise $\boldsymbol{\epsilon}^\mathcal{S}_\theta$ and image component $\boldsymbol{C}^\mathcal{S}_{t_{\boldsymbol{\theta}}}$. This process can be formulated by:
\begin{equation}
\begin{aligned}
    & \\
    &\boldsymbol{C}^\mathcal{S}_{t_{\boldsymbol{\theta}}}, \boldsymbol{\epsilon}^\mathcal{S}_{\boldsymbol{\theta}} = \boldsymbol{\mathrm{Net}}^\mathcal{S}_{\boldsymbol{\theta}}(\boldsymbol{\mathrm{x}}^\mathcal{S}_t, t),
\end{aligned}
\end{equation}
where $\boldsymbol{\theta}$ represents the parameter of domain $\mathcal{S}$ denoising U-Net. Note that we perform the same implementation in domain $\mathcal{T}$.

After obtaining the image components $\boldsymbol{C}^\mathcal{S}_{t_{\boldsymbol{\theta}}}$ and $\boldsymbol{C}^\mathcal{T}_{t_{\boldsymbol{\theta}}}$ at time $t$, we propose a cycle translator to conduct the cycle translation process as shown in the middle of Fig.~\ref{fig_architecture}. 
The cycle translator includes two time-dependent translation networks $F_{\boldsymbol{\psi}}$ and $G_{\boldsymbol{\phi}}$, which represents the mapping $\mathcal{T}\to\mathcal{S}$ and $\mathcal{S}\to\mathcal{T}$ respectively.  
Given the image component $\boldsymbol{C}^\mathcal{S}_{t_{\boldsymbol{\theta}}}$ from domain $\mathcal{S}$, the time-dependent translation network $G_{\boldsymbol{\phi}}$ first translate it into domain $\mathcal{T}$, and then the translated image component is projected back to domain $\mathcal{S}$, resulting in a cycle translation process. 
We formulate this process by the following equation:
\begin{equation}
    \boldsymbol{\hat{C}}^\mathcal{S}_{t_{\boldsymbol{\theta}}} = F_{\boldsymbol{\psi}}(G_{\boldsymbol{\phi}}(\boldsymbol{C}^\mathcal{S}_{t_{\boldsymbol{\theta}}}, t), t),
\end{equation}
In a similar way, the cycle translation of $\mathcal{T}\to\mathcal{S}\to\mathcal{T}$ can be written as:
\begin{equation}
    \boldsymbol{\hat{C}}^\mathcal{T}_{t_{\boldsymbol{\theta}}} = G_{\boldsymbol{\phi}}(F_{\boldsymbol{\psi}}(\boldsymbol{C}^\mathcal{T}_{t_{\boldsymbol{\theta}}}, t), t).
\end{equation}

\subsubsection{Time-dependent translation network}\label{3.1.2}
Since we jointly learn the translation and the diffusion processes, we need to view the translation process as a multi-step mapping and train the cycle translator at each time $t$. To this end, we introduce a time-dependent translation network that fuses the time information into the feature of the image component.
Concretely, the time-dependent translation network comprises a time-attention block and a ResNet\cite{he2016deep}-based encoder-decoder architecture. 
In the time-attention block, the time step $t$ is first encoded via an MLP to obtain a time embedding $\dot{t}$. This embedding is then fused with the estimated image component $\boldsymbol{C}^\mathcal{S}_{t_{\boldsymbol{\theta}}}$ using FiLM~\cite{perez2018film} mechanism, resulting in a fused feature.
Then, a self-attention layer is adopted to enhance the fused feature. The process is represented as:
\begin{equation}
\begin{aligned}
    &\dot{t} = \mathbf{MLP}(t), \\
    {\boldsymbol{\bar{C}}^\mathcal{S}_{t_{\boldsymbol{\theta}}}} &= \mathbf{Resnet Block}({\boldsymbol{{C}}^\mathcal{S}_{t_{\boldsymbol{\theta}}}}, \dot{t}), \\
    {\boldsymbol{\tilde{C}}^\mathcal{S}_{t_{\boldsymbol{\theta}}}} &= \mathbf{MHSA}({\boldsymbol{\bar{C}}^\mathcal{S}_{t_{\boldsymbol{\theta}}}}),
\end{aligned}
\end{equation}
where $\mathbf{MLP}(\cdot)$ and $\mathbf{ResnetBlock}(\cdot)$, and $\mathbf{MHSA}(\cdot)$ mean the MLP layer, Resnet block, and multi-head self-attention layer. ${\boldsymbol{\bar{C}}^\mathcal{S}_{t_{\boldsymbol{\theta}}}}$ and ${\boldsymbol{\tilde{C}}^\mathcal{S}_{t_{\boldsymbol{\theta}}}}$ denote the intermediate output after Resnet block and the enhanced feature after self-attention respectively.
Next, the enhanced feature ${\boldsymbol{\tilde{C}}^\mathcal{S}_{t_{\boldsymbol{\theta}}}}$ is fed into the encoder-decoder architecture that is composed of stacked Resnet blocks, producing the translated feature. 
Finally, a $7\times 7$ convolution is applied to the translated feature, resulting in the translated image component.

\begin{figure*}[p]
    \centering
    \includegraphics[width=1.02\textwidth]{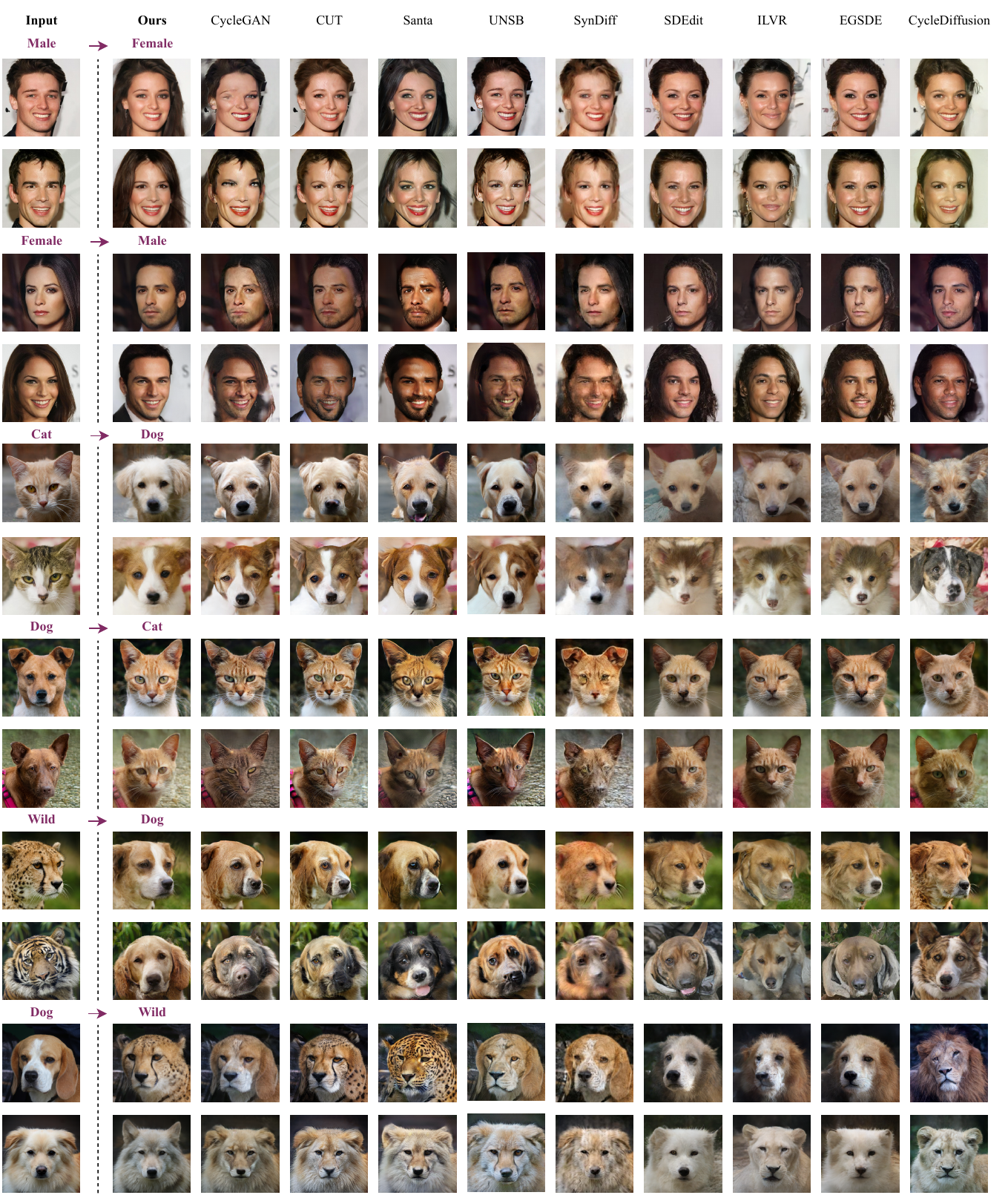}
    \caption{Qualitative comparisons on RGB $\leftrightarrow$ RGB tasks with state-of-the-art methods. CycleDiff could achieve superior visual results for both realism and faithfulness across all tasks. For example, in the fourth row, our method effectively retains the features that are independent of the domain, such as the white ground, while eliminating those that are specific to the domain, such as the shape of the eyebrows and mouth.}
    \label{fig_all_public}
\end{figure*}

\begin{figure*}[t]
    \centering
    \includegraphics[width=0.98\textwidth]{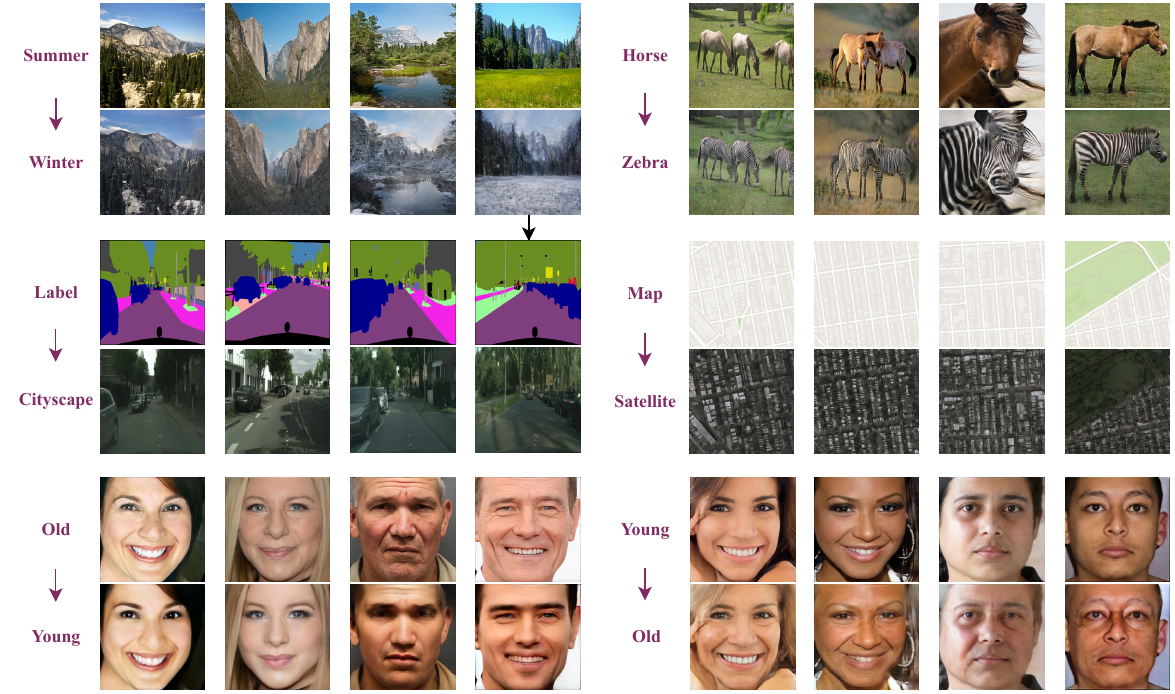}
    \caption{More visual results on additional datasets of CycleDiff. Our method could produce high fidelity results both on time-varying datasets and challenging artificial domain data.}
    \label{fig_additional_dataset}
\end{figure*}

\subsection{Training}\label{3.2}
The training objective of CycleDiff consists of two main components: the diffusion model loss and the cycle translator loss. The diffusion loss is used to supervise the denoising network and extract the image component. The cycle translator loss includes three elements. Firstly, the adversarial loss and discriminator contrastive loss (DCL) ensure that the distributions of translated image components closely match those of the target domain. Secondly, the cycle consistency loss and perceptual loss encourage the mapping function to preserve crucial features of the source image components. Thirdly, the identity loss aims to further enhance the quality of the translated image components while preserving maximum background information. The detailed training process is illustrated in Alg.~\ref{alg:1}.

\subsubsection{Diffusion model loss}
For the training of the decoupled diffusion model, we supervise the $\boldsymbol{\epsilon}^\mathcal{S}$ and $\boldsymbol{\mathrm{x}}^\mathcal{S}_0$ simultaneously, thus, the training objective of the diffusion model $\mathcal{L}_{dm}$ is represented as:
\begin{equation}
    \mathcal{L}_{dm} = \min\limits_{\boldsymbol{\theta}} \mathbb{E}_{q(\boldsymbol{\mathrm{x}}^\mathcal{S}_{0})} \mathbb{E}_{q(\boldsymbol{\epsilon})} [\Vert \boldsymbol{C}^\mathcal{S}_{t_{\boldsymbol{\theta}}} - \boldsymbol{C}^\mathcal{S}_{t} \Vert^{2} + \Vert \boldsymbol{\epsilon}^\mathcal{S}_{\boldsymbol{\theta}}-\boldsymbol{\epsilon}^\mathcal{S}\Vert^{2}],
\label{eq_loss_ddm}
\end{equation}
where $\boldsymbol{\theta}$ represents the parameter of denoising network, $\boldsymbol{C}^\mathcal{S}_{t_{\boldsymbol{\theta}}}$ and $\boldsymbol{\epsilon}^\mathcal{S}_{\boldsymbol{\theta}}$ denote the output of the diffusion model. Note that the domain $\mathcal{T}$ employs the same loss.

\subsubsection{Cycle translator loss}

\input{tables/alg_1}

\textbf{Adversarial Loss}
To train the cycle translator, we employ the classical adversarial loss and two additional discriminators $D^\mathcal{S}_{\boldsymbol{\psi}}$ and $D^\mathcal{T}_{\boldsymbol{\phi}}$ to restraint mapping function $F_{\boldsymbol{\psi}}$ and $G_{\boldsymbol{\phi}}$. For the mapping function $G_{\boldsymbol{\phi}}: \mathcal{S}\to\mathcal{T}$ and its corresponding discriminator $D^\mathcal{T}_{\boldsymbol{\phi}}$, the adversarial loss can be expressed as follows:
\begin{equation}
\begin{aligned}
    \mathcal{L}^\mathcal{S}_{adv} &= \mathbb{E}_{\boldsymbol{C}^\mathcal{T}_{t_{\boldsymbol{\theta}}} \sim p_{data}(\boldsymbol{C}^\mathcal{T}_{t_{\boldsymbol{\theta}}})}[\mathrm{log} D^\mathcal{T}_{\boldsymbol{\phi}}(\boldsymbol{C}^\mathcal{T}_{t_{\boldsymbol{\theta}}})] \\
    & + \mathbb{E}_{\boldsymbol{C}^\mathcal{S}_{t_{\boldsymbol{\theta}}} \sim p_{data}(\boldsymbol{C}^\mathcal{S}_{t_{\boldsymbol{\theta}}})}[\mathrm{log} (1 - D^\mathcal{T}_{\boldsymbol{\phi}}(G_{\boldsymbol{\phi}}(\boldsymbol{C}^\mathcal{S}_{t_{\boldsymbol{\theta}}}, t)))],
\end{aligned}
\end{equation}
$G_{\boldsymbol{\phi}}$ tries to generate image components that are similar to the image components from domain $\mathcal{T}$, while $D^\mathcal{T}_{\boldsymbol{\phi}}$ aims to distinguish between the translated image components $G_{\boldsymbol{\phi}}(\boldsymbol{C}^\mathcal{S}_{t_{\boldsymbol{\theta}}}, t)$ and real image components $\boldsymbol{C}^\mathcal{T}_{t_{\boldsymbol{\theta}}}$. The goal of $G_{\boldsymbol{\phi}}$ is to minimize the objective while competing with an adversary $D_{\boldsymbol{\phi}}$, which attempts to maximize it, i.e. $\mathrm{min}_{G_{\boldsymbol{\phi}}} \mathrm{max}_{D^\mathcal{T}_{\boldsymbol{\phi}}} \mathcal{L}^\mathcal{S}_{adv}(G_{\boldsymbol{\phi}}, D^\mathcal{T}_{\boldsymbol{\phi}}, \boldsymbol{C}^\mathcal{S}_{t_{\boldsymbol{\theta}}}, \boldsymbol{C}^\mathcal{T}_{t_{\boldsymbol{\theta}}})$. Similarly, the objective for mapping function $F_{\boldsymbol{\psi}}: \mathcal{T} \to \mathcal{S}$ and its discriminator $D^\mathcal{S}_{\boldsymbol{\psi}}$ as well: i.e. $\mathrm{min}_{F_{\boldsymbol{\psi}}} \mathrm{max}_{D^\mathcal{S}_{\boldsymbol{\psi}}} \mathcal{L}^\mathcal{T}_{adv}(F_{\boldsymbol{\psi}}, D^\mathcal{S}_{\boldsymbol{\psi}}, \boldsymbol{C}^\mathcal{T}_{t_{\boldsymbol{\theta}}}, \boldsymbol{C}^\mathcal{S}_{t_{\boldsymbol{\theta}}})$. In summary, we can define the adversarial loss as:
\begin{equation}
\begin{aligned}
    \mathcal{L}_{adv} &= \mathcal{L}^\mathcal{S}_{adv}(G_{\boldsymbol{\phi}}, D^\mathcal{T}_{\boldsymbol{\phi}}, \boldsymbol{C}^\mathcal{S}_{t_{\boldsymbol{\theta}}}, \boldsymbol{C}^\mathcal{T}_{t_{\boldsymbol{\theta}}}) \\
    &+ \mathcal{L}^\mathcal{T}_{adv}(F_{\boldsymbol{\psi}}, D^\mathcal{S}_{\boldsymbol{\psi}}, \boldsymbol{C}^\mathcal{T}_{t_{\boldsymbol{\theta}}}, \boldsymbol{C}^\mathcal{S}_{t_{\boldsymbol{\theta}}}).
\end{aligned}
\end{equation}

\input{tables/alg_2}

\textbf{DCL loss}
To stabilize the training procedure of the generator and make full use of the discriminator output, we introduce Discriminator Contrastive Loss (DCL) instead of just mapping the input image to a probability scalar via the discriminator. We first reshape the output of the discriminator $\mathbf{\hat{v}}^\mathcal{T}$ to $N$-dimensional feature vectors (here, we set $N=7$). Then, we normalize each vector to prevent from space collapsing and get $\mathbf{\dot{\hat{v}}}^\mathcal{T}$. The DCL loss can be formulated as follows:
\begin{equation}
\begin{aligned}
    &\mathcal{L}^\mathcal{T}_{dcl}(\mathbf{\dot{v}}^\mathcal{T}_{i}, \mathbf{\dot{v}}^\mathcal{T}, \mathbf{\dot{v}}^\mathcal{T}_{i-}) = - \frac{1}{\left | \mathbf{\dot{v}}^\mathcal{T}_{i-} \right | } \sum_{\mathbf{\dot{v}}^\mathcal{T}_{j} \in \mathbf{\dot{v}}^\mathcal{T}_{i-}} \mathrm{log} \times \\
    & \frac{\mathrm{exp}(\mathbf{\dot{v}}^\mathcal{T}_{i} \cdot \mathbf{\dot{v}}^\mathcal{T}_{j} / \tau)}{\sum_{\mathbf{\dot{\hat{v}}}^\mathcal{T}_{k} \in \mathbf{\dot{\hat{v}}}^\mathcal{T}} \mathrm{exp}(\mathbf{\dot{v}}^\mathcal{T}_{i} \cdot \mathbf{\dot{\hat{v}}}^\mathcal{T}_{k} / \tau) + \sum_{\mathbf{\dot{v}}^\mathcal{T}_{k} \in \mathbf{\dot{v}}^\mathcal{T}_{i-}} \mathrm{exp}(\mathbf{\dot{v}}^\mathcal{T}_{i} \cdot \mathbf{\dot{v}}^\mathcal{T}_{k} / \tau)},
\end{aligned}
\end{equation}
where $\mathbf{\dot{v}}^\mathcal{T}_{i-}$ denotes those feature vectors are in $\mathbf{\dot{v}}^\mathcal{T}$ but not in $\mathbf{\dot{v}}^\mathcal{T}_{i}$, i.e., $\mathbf{\dot{v}}^\mathcal{T}_{i-} = \mathbf{\dot{v}}^\mathcal{T} \backslash \mathbf{\dot{v}}^\mathcal{T}_{i}$. $\tau=0.1$ is the temperature, which scales the distance between different feature vectors. The DCL loss for $\mathcal{L}^\mathcal{S}_{dcl}(\mathbf{\dot{v}}^\mathcal{S}_{i}, \mathbf{\dot{v}}^\mathcal{S}, \mathbf{\dot{v}}^\mathcal{S}_{i-})$ is calculated in a similar manner.

\textbf{Cycle Consistency Loss}
Despite the $L_{adv}$ could ensure the translated image component $\boldsymbol{\hat{C}}^\mathcal{T}_{t_{\boldsymbol{\theta}}}$ to be in the correct domain $\mathcal{T}$. However, this cannot encourage the translated image component $\boldsymbol{\hat{C}}^\mathcal{T}_{t_{\boldsymbol{\theta}}}$ to be similar to the source image component $\boldsymbol{C}^\mathcal{S}_{t_{\boldsymbol{\theta}}}$. To preserve important features of the source component, we introduce the cycle consistency loss to maintain the structure similarity. Specifically, for each image component $\boldsymbol{C}^\mathcal{S}_{t_{\boldsymbol{\theta}}}$ from $\mathcal{S}$, the cycle translation encourages the source image component $\boldsymbol{C}^\mathcal{S}_{t_{\boldsymbol{\theta}}}$ to translate back to the source image component, i.e., $\boldsymbol{C}^\mathcal{S}_{t_{\boldsymbol{\theta}}} \to G_{\boldsymbol{\phi}}(\boldsymbol{C}^\mathcal{S}_{t_{\boldsymbol{\theta}}}, t) \to F_{\boldsymbol{\psi}}(G_{\boldsymbol{\phi}}(\boldsymbol{C}^\mathcal{S}_{t_{\boldsymbol{\theta}}},t ), t) \approx \boldsymbol{C}^\mathcal{S}_{t_{\boldsymbol{\theta}}}$.
The cycle consistency loss can be formulated as follows:
\begin{equation}
\begin{aligned}
    \mathcal{L}_{cyc} &= \mathbb{E}_{\boldsymbol{C}^\mathcal{S}_{t_{\boldsymbol{\theta}}} \sim p_{data}(\boldsymbol{C}^\mathcal{S}_{t_{\boldsymbol{\theta}}})}[\left \| F_{\boldsymbol{\psi}}(G_{\boldsymbol{\phi}}(\boldsymbol{C}^\mathcal{S}_{t_{\boldsymbol{\theta}}}, t), t) -\boldsymbol{C}^\mathcal{S}_{t_{\boldsymbol{\theta}}} \right \|_1 ] \\
    &+ \mathbb{E}_{\boldsymbol{C}^\mathcal{T}_{t_{\boldsymbol{\theta}}} \sim p_{data}(\boldsymbol{C}^\mathcal{T}_{t_{\boldsymbol{\theta}}})}[\left \| G_{\boldsymbol{\phi}}(F_{\boldsymbol{\psi}}(\boldsymbol{C}^\mathcal{T}_{t_{\boldsymbol{\theta}}}, t), t) -\boldsymbol{C}^\mathcal{T}_{t_{\boldsymbol{\theta}}} \right \|_1 ].
\end{aligned}
\end{equation}

\textbf{Perceptual Loss}
Considering the $\mathcal{L}_{cyc}$ is not enough to recover all textural and structural information, we incorporate the perceptual loss \cite{johnson2016perceptual} to preserve the source image component structure. We combine the loss of the high and low-level feature by feeding the source image component $\boldsymbol{C}^\mathcal{S}_{t_{\boldsymbol{\theta}}}$ and translated image component $F_{\boldsymbol{\psi}}(G_{\boldsymbol{\phi}}(\boldsymbol{C}^\mathcal{S}_{t_{\boldsymbol{\theta}}}))$ into VGG16 \cite{vgg16} architecture. The perceptual loss is as follows:
\begin{equation}
\begin{aligned}
    &\mathcal{L}_{lps} = \mathbb{E}_{\boldsymbol{C}^\mathcal{S}_{t_{\boldsymbol{\theta}}} \sim p_{data}(\boldsymbol{C}^\mathcal{S}_{t_{\boldsymbol{\theta}}})}[\left \| \varphi(F_{\boldsymbol{\psi}}(G_{\boldsymbol{\phi}}(\boldsymbol{C}^\mathcal{S}_{t_{\boldsymbol{\theta}}}, t), t)) -\varphi(\boldsymbol{C}^\mathcal{S}_{t_{\boldsymbol{\theta}}}) \right \|_2 ] \\
    &+ \mathbb{E}_{\boldsymbol{C}^\mathcal{T}_{t_{\boldsymbol{\theta}}} \sim p_{data}(\boldsymbol{C}^\mathcal{T}_{t_{\boldsymbol{\theta}}})}[\left \| \varphi(G_{\boldsymbol{\phi}}(F_{\boldsymbol{\psi}}(\boldsymbol{C}^\mathcal{T}_{t_{\boldsymbol{\theta}}}, t), t)) -\varphi(\boldsymbol{C}^\mathcal{T}_{t_{\boldsymbol{\theta}}}) \right \|_2 ],
\end{aligned}
\end{equation}
where $\varphi(\cdot)$ is the VGG16 feature extractor from 1$\sim$5-th convolutional layers.

\textbf{Identity Loss}
To ensure approximate identity mappings when the target domain image components are fed into the generator, we introduce identity loss to maintain this mapping. 
Formally, the identity loss is defined as follows:
\begin{equation}
\begin{aligned}
    \mathcal{L}_{idt} &= \mathbb{E}_{\boldsymbol{C}^\mathcal{S}_{t_{\boldsymbol{\theta}}} \sim p_{data}(\boldsymbol{C}^\mathcal{S}_{t_{\boldsymbol{\theta}}})}[\left \| F_{\boldsymbol{\psi}}(\boldsymbol{C}^\mathcal{S}_{t_{\boldsymbol{\theta}}}, t) -\boldsymbol{C}^\mathcal{S}_{t_{\boldsymbol{\theta}}} \right \|_1 ] \\
    &+ \mathbb{E}_{\boldsymbol{C}^\mathcal{T}_{t_{\boldsymbol{\theta}}} \sim p_{data}(\boldsymbol{C}^\mathcal{T}_{t_{\boldsymbol{\theta}}})}[\left \| G_{\boldsymbol{\phi}}(\boldsymbol{C}^\mathcal{T}_{t_{\boldsymbol{\theta}}}, t) -\boldsymbol{C}^\mathcal{T}_{t_{\boldsymbol{\theta}}} \right \|_1 ].
\end{aligned}
\end{equation}

\subsubsection{Full Optimization Objective of CycleDiff}
The full optimization objective is composed of the diffusion loss and translation loss:
\begin{equation}
\begin{aligned}
    \mathcal{L} &= \lambda_1 \mathcal{L}_{dm} + \mathcal{L}_{tra} \\
    &= \lambda_1 \mathcal{L}_{dm} + \lambda_2 \mathcal{L}_{adv} + \lambda_3 \mathcal{L}_{cyc} \\ &+ \lambda_4 \mathcal{L}_{idt} + \lambda_5 \mathcal{L}_{lps} + \lambda_6 \mathcal{L}_{dcl},
\end{aligned}
\end{equation}
where $\lambda_1, \lambda_2, \lambda_3, \lambda_4, \lambda_5, \lambda_6$ are scale values, controlling the weight of different losses. Compared with CycleGAN, we conduct an ablation experiment on the newly added loss of $\mathcal{L}_{lps}$ and $\mathcal{L}_{dcl}$ in Sec.~\ref{ablation}.

\input{tables/tab_all}
\input{tables/tab_add}

\subsection{Inference}\label{3.3}
Since we learn the translation process as a multi-step mapping, we need to call the translation network in the inference stage iteratively.
Specifically, given an image from domain $S$,  we first utilize the reverse diffusion sampling \cite{wu2023cyclediffusion} to obtain its image component $\boldsymbol{C}^\mathcal{S}_{t_\theta}$ at each denoising step. At the same time, we forward $\boldsymbol{C}^\mathcal{S}_{t_\theta}$ into the translation network $G_{\boldsymbol{\phi}}$ together with the time step $t$ to produce $\boldsymbol{\hat{C}}^\mathcal{T}_{t_\theta}$. To generate the corresponding image of domain $\mathcal{T}$, we start from the normal distribution and conduct the iterative denoising process using the diffusion model of domain $\mathcal{T}$. 
Importantly, we replace the image component of domain $\mathcal{T}$ with the translated image component $\boldsymbol{C}^\mathcal{S}_{t_\theta}$ at each time $t$. In this way, we can utilize the diffusion model of domain $\mathcal{T}$ to generate images corresponding to domain $\mathcal{S}$ and vice versa. 
The detailed sampling process is shown in Alg.~\ref{alg:2}.

\subsection{Implementation Details}\label{3.4}

\textbf{Network architecture}
As shown at the bottom of Fig.~\ref{fig_architecture}, the proposed time-dependent translation network consists of the time-attention block and an encoder-decoder architecture. More specifically, the time-attention block includes an MLP layer, a ResNet block, and a self-attention layer, while the encoder-decoder architecture consists of three down-sampling layers, twelve residual blocks, and three up-sampling layers. The ablation on variants of translation network architecture in Sec.~\ref{ablation}. 
We follow \cite{rombach2022high} to conduct the diffusion process in the latent space.
For the design of the discriminator, we adopt the same discriminator framework as PatchGANs \cite{isola2017image}.
The architecture of the diffusion model is based on UNet~\cite{song2020improved}. Inspired by~\cite{huangddm}, our denoising UNet needs to output both the image component and the estimated noise. To accommodate this, the UNet comprises two decoders for predicting the image component $\boldsymbol{C}$ and the noise component $\boldsymbol{\epsilon}$ respectively.

\textbf{Training details} We perform all experiments on a single NVIDIA A100 GPU. We select the least-square loss \cite{mao2017least} as the adversarial loss.
For training of the CycleDiff, the training process contains 100000 iterations. We warm up the diffusion models from scratch for the first 50000 iterations, then the diffusion and translation models are optimized together until the training stops.
For training of the diffusion model, we employ an AdamW optimizer with a decaying learning rate (from $1e^{-4}$ to $1e^{-5}$).
For the joint training of the cycle translator, we apply Adam optimizer with an initial learning rate of $2e^{-4}$. The batch size is set to 24 for all experiments. 
We utilize the exponential moving average (EMA) to stabilize the performance of models during training. 
For the perceptual loss, we adopt the commonly used VGG16 with the official pretrained weights. We calculate the L2 loss between the features extracted from the 1$\sim$5-th convolutional layers of the translated and source image components. We then average them to obtain the final perceptual loss.
We set $\lambda_1 = 5e^{-2}, \lambda_2 = 1, \lambda_3 = 10, \lambda_4 = 5, \lambda_5 = 0.5, \lambda_6 = 0.02$ as the standard setting for all experiments. In the denoising process, we implement 100 diffusion steps for RGB $\leftrightarrow$ RGB tasks, aligning with the SDDM method, and extend this to 200 steps for cross-modality tasks. The ablation study on the effect of additional denoising steps can be seen in Sec.~\ref{ablation}.

%% file: tables/alg_1.tex
\begin{algorithm}[t]
    \caption{Training algorithm of CycleDiff.}
    \begin{algorithmic}[1]  
    \State \textbf{Initialize} $i=0, N=num\_iters, lr$; denoising network parameters: $\boldsymbol{\theta}$; cycle translator parameters: $\boldsymbol{\phi}$, $\boldsymbol{\psi}$; diffusion model $\boldsymbol{\mathrm{Net}}^\mathcal{S}_{\boldsymbol{\theta}}$ of domain $\mathcal{S}$ and $\boldsymbol{\mathrm{Net}}^\mathcal{T}_{\boldsymbol{\theta}}$ of domain $\mathcal{T}$; translation network $G_{\boldsymbol{\phi}}$ for $\mathcal{S} \to \mathcal{T}$ and $F_{\boldsymbol{\psi}}$ for $\mathcal{T} \to \mathcal{S}$;
    \While {$i<N$}
        \State $x^\mathcal{S}_0 \in \mathcal{S}, x^\mathcal{T}_0 \in \mathcal{T}$;
        \State $t\sim Uniform(0, 1), \boldsymbol{\epsilon}^\mathcal{S} \sim \mathcal{N}(\mathbf{0}, \mathbf{I}),
        \boldsymbol{\epsilon}^\mathcal{T} \sim \mathcal{N}(\mathbf{0}, \mathbf{I})$;
        \State $\mathbf{x}^\mathcal{S}_t = \mathbf{x}^\mathcal{S}_0 + \int_0^t \boldsymbol{C}^\mathcal{S}_t \mathrm{d}t + t\boldsymbol{\epsilon}^\mathcal{S}$;
        \State $\mathbf{x}^\mathcal{T}_t = \mathbf{x}^\mathcal{T}_0 + \int_0^t \boldsymbol{C}^\mathcal{T}_t \mathrm{d}t + t\boldsymbol{\epsilon}^\mathcal{T}$;
        \State $\boldsymbol{C}^\mathcal{S}_{t_{\boldsymbol{\theta}}}, \boldsymbol{\epsilon}^\mathcal{S}_{\boldsymbol{\theta}} = \boldsymbol{\mathrm{Net}}^\mathcal{S}_{\boldsymbol{\theta}}(\mathbf{x}^\mathcal{S}_{t}, t), \boldsymbol{C}^\mathcal{T}_{t_{\boldsymbol{\theta}}}, \boldsymbol{\epsilon}^\mathcal{T}_{\boldsymbol{\theta}} = \boldsymbol{\mathrm{Net}}^\mathcal{T}_{\boldsymbol{\theta}}(\mathbf{x}^\mathcal{T}_{t}, t)$;
        \State $\boldsymbol{\hat{C}}^\mathcal{S}_{t_{\boldsymbol{\theta}}} = F_{\boldsymbol{\psi}}(G_{\boldsymbol{\phi}}(\boldsymbol{C}^\mathcal{S}_{t_{\boldsymbol{\theta}}}, t), t), \boldsymbol{\hat{C}}^\mathcal{T}_{t_{\boldsymbol{\theta}}} = G_{\boldsymbol{\phi}}(F_{\boldsymbol{\psi}}(\boldsymbol{C}^\mathcal{T}_{t_{\boldsymbol{\theta}}}, t), t)$;
        \State Calculate $\mathcal{L}_{dm}$ and $\mathcal{L}_{tra}$;
        \State $\boldsymbol{\theta, \phi, \psi} \leftarrow lr*\nabla_{\boldsymbol{\theta, \phi, \psi}} (\mathcal{L}_{dm} + \mathcal{L}_{tra}$);
        \State $i = i+1$;
    \EndWhile; \\
    \Return $\boldsymbol{\theta}, \boldsymbol{\phi}, \boldsymbol{\psi}$;  
    \end{algorithmic}  
    \label{alg:1}
\end{algorithm}

%% file: tables/alg_2.tex
\begin{algorithm}[t]
    \caption{Inference algorithm of CycleDiff.}
    \begin{algorithmic}[1]  
    \State \textbf{Initialize} $t=0, k=0, N=num\_steps, s=1/N, x^\mathcal{S}_0 \in \mathcal{S}, \boldsymbol{\epsilon}^\mathcal{S} \sim \mathcal{N}(\mathbf{0}, \mathbf{I}), \boldsymbol{\mathrm{Net}}^\mathcal{S}_{\boldsymbol{\theta}}, \boldsymbol{\mathrm{Net}}^\mathcal{T}_{\boldsymbol{\theta}}, G_{\boldsymbol{\phi}}, \boldsymbol{C}_{list}=[]$;
    \State $\mathbf{z}^\mathcal{S}_t = \mathbf{z}^\mathcal{S}_0 + \int_0^t \boldsymbol{C}^\mathcal{S}_t \mathrm{d}t + t\boldsymbol{\epsilon}^\mathcal{S}$;
    \State $\boldsymbol{C}^\mathcal{S}_{t_{\boldsymbol{\theta}}}, \boldsymbol{\epsilon}^\mathcal{S}_{\boldsymbol{\theta}} = \boldsymbol{\mathrm{Net}}^\mathcal{S}_{\boldsymbol{\theta}}(\mathbf{z}^\mathcal{S}_{t}, t), \boldsymbol{C}^\mathcal{T}_{t_{\boldsymbol{\theta}}}$;
    \State $t=t+s$;
    \State $\boldsymbol{C}_{list}.\mathbf{append}(\boldsymbol{C}^\mathcal{S}_{t_{\boldsymbol{\theta}}})$;
    \While {$t<=1$}
        \State $\mathbf{z}^\mathcal{S}_t = \mathbf{z}^\mathcal{S}_0 + \int_0^t \boldsymbol{C}^\mathcal{S}_{t_{\boldsymbol{\theta}}} \mathrm{d}t + t\boldsymbol{\epsilon}^\mathcal{S}_{\boldsymbol{\theta}}$;
        \State $\boldsymbol{C}^\mathcal{S}_{t_{\boldsymbol{\theta}}}, \boldsymbol{\epsilon}^\mathcal{S}_{\boldsymbol{\theta}} = \boldsymbol{\mathrm{Net}}^\mathcal{S}_{\boldsymbol{\theta}}(\mathbf{z}^\mathcal{S}_{t}, t)$;
        \State $\boldsymbol{\hat{C}}^\mathcal{T}_{t_{\boldsymbol{\theta}}} = G_{\boldsymbol{\phi}}(\boldsymbol{C}^\mathcal{S}_{t_{\boldsymbol{\theta}}}, t)$;
        \State $\boldsymbol{C}_{list}.\mathbf{append}(\boldsymbol{\hat{C}}^\mathcal{T}_{t_{\boldsymbol{\theta}}})$;
        \State $t=t+s$;
    \EndWhile;
    \State $\mathbf{x}^\mathcal{T}_t \sim \mathcal{N}(\mathbf{0}, \mathbf{I})$;
    \While {$t>0$}
        \State $\boldsymbol{C}^\mathcal{T}_{t_{\boldsymbol{\theta}}}, \boldsymbol{\epsilon}^\mathcal{T}_{\boldsymbol{\theta}} = \boldsymbol{\mathrm{Net}}^\mathcal{T}_{\boldsymbol{\theta}}(\mathbf{x}^\mathcal{T}_{t}, t)$;
        \State $\boldsymbol{C}^\mathcal{S}_{t_{\boldsymbol{\theta}}} = \boldsymbol{C}_{list}.\mathbf{pop}()$;
        \State $\boldsymbol{\mathrm{d}} = \boldsymbol{C}^\mathcal{S}_{t_{\boldsymbol{\theta}}} + \boldsymbol{\epsilon}^\mathcal{T}_{\boldsymbol{\theta}}$;
        \State $\mathbf{x}^\mathcal{T}_{t-s} = \mathbf{x}^\mathcal{T}_{t} - \boldsymbol{\mathrm{d}}(t-s)$;
        \State $\mathbf{x}^\mathcal{T}_t = \mathbf{x}^\mathcal{T}_{t-s}$;
        \State $t=t-s$;
    \EndWhile;\\
    \Return $\mathbf{x}^\mathcal{T}_t$;
    \end{algorithmic}  
    \label{alg:2}
\end{algorithm}

%% file: tables/tab_all.tex
\begin{table*}[ht]
\begin{center}
\caption{
{Quantitative comparison of unpaired image-to-image translation methods. The best results are shown in \textbf{bold}, and the second-best results are \underline{underlined}. Note that the KID metric is multiplied by 100.}
}\label{tab_all_public}
\vspace{0.1cm}
\normalsize
\setlength{\tabcolsep}{6pt}
\renewcommand{\arraystretch}{0.925}
{
\begin{tabular}{lcccccc}
\hline
Model                                     & FID $\downarrow$       & KID $\downarrow$ & SSIM $\uparrow$         & FID $\downarrow$       & KID $\downarrow$ & SSIM $\uparrow$         \\ \hline
\multicolumn{1}{c}{}                      & \multicolumn{3}{c}{Cat $\to$ Dog}                                   & \multicolumn{3}{c}{Dog $\to$ Cat}                                   \\ \hline
CycleGAN\cite{zhu2017cyclegan}            & 85.9                   & -                & -                       & 107.7                  & -                & -                       \\
MUNIT\cite{huang2018munit}                & 104.4                  & -                & -                       & -                      & -                & -                       \\
DRIT\cite{lee2018drit}                    & 123.4                  & -                & -                       & -                      & -                & -                       \\
Distance\cite{benaim2017distance}         & 155.3                  & -                & -                       & -                      & -                & -                       \\
SelfDistance\cite{benaim2017distance}     & 144.4                  & -                & -                       & -                      & -                & -                       \\
GCGAN\cite{fu2019gcgan}                   & 96.6                   & -                & -                       & -                      & -                & -                       \\
LSeSim\cite{zheng2021lsesim}              & 72.8                   & -                & -                       & -                      & -                & -                       \\
ITTR (CUT)\cite{zheng2022ittr}            & 68.6                   & -                & -                       & -                      & -                & -                       \\
StarGAN v2\cite{choi2020starganv2}        & 54.88 ± 1.01           & -                & 0.27 ± 0.003            & -                      & -                & -                       \\
CUT\cite{park2020cut}                     & 76.21                  & 4.287                & \textbf{0.601}          & -                      & -                & -                       \\
Santa\cite{xie2023unpaired}                     & 52.1                  & 2.773                & 0.492          & 46.06                      & 1.678                & 0.415                       \\ \hline
UNSB\cite{UNSB}                           & 79.69           & 3.936                & 0.543           & \underline{48.31} & \underline{1.814}                & \textbf{0.614}           \\
SynDiff\cite{ozbey2023unsupervised}   & 148.79                & 9.926              & 0.514                 & 171.52               & 12.051                  & 0.579      \\
ILVR\cite{ILVR}                           & 74.37 ± 1.55           & 2.054 ± 0.231                & 0.363 ± 0.001           & 54.52 ± 1.34 & 3.298 ± 0.196                & 0.392 ± 0.001           \\
SDEdit\cite{SDEdit}                       & 74.17 ± 1.01           & 2.357 ± 0.156                & 0.423 ± 0.001           & 59.44 ± 1.01           & 3.736 ± 0.134                & 0.425 ± 0.001 \\
EGSDE\cite{zhao2022egsde}                 & 70.16 ± 1.03           & \underline{1.982 ± 0.147}                & 0.411 ± 0.001           & 62.34 ± 1.54           & 3.191 ± 0.167                & 0.402 ± 0.001           \\
SDDM\cite{sun2023sddm}                    & 62.29 ± 0.63           & -                & 0.422 ± 0.001           & -                      & -                & -                       \\
CycleDiffusion\cite{wu2023cyclediffusion} & \underline{58.87}        & 2.140                & \textbf{0.557}          & 58.80                  & 1.510                & 0.403                   \\
CycleDiff (Ours)                          & \textbf{49.78 ± 0.43}  & \textbf{1.487 ± 0.105}                & \underline{0.552 ± 0.001} & \textbf{26.45 ± 0.36}  & \textbf{0.847 ± 0.096}                & \underline{0.585 ± 0.001}  \\ \hline
\multicolumn{1}{c}{}                      & \multicolumn{3}{c}{Wild $\to$ Dog}                                  & \multicolumn{3}{c}{Dog $\to$ Wild}                                  \\ \hline
CycleGAN\cite{zhu2017cyclegan}            & 83.95                  & -                & -                       & -                      & -                & -                       \\
CUT\cite{park2020cut}                     & 92.94                  & 4.807                & \textbf{0.592}          & -                      & -                & -                       \\ 
Santa\cite{xie2023unpaired}                     & 74.93                  & 3.161                & 0.453          & 43.44                      & 1.714                & 0.439                       \\ \hline
UNSB\cite{UNSB}                           & 86.72           & 4.018                & \underline{0.524}           & \underline{59.58} & 2.470                & \underline{0.582}           \\ 
SynDiff\cite{ozbey2023unsupervised}   & 133.08               & 6.653              & 0.201                 & 156.53               & 8.732                  & 0.168      \\
ILVR\cite{ILVR}                           & 75.33 ± 1.22           & 2.766 ± 0.180                & 0.287 ± 0.001           & 63.15 ± 1.14 & 2.918 ± 0.175                & 0.353 ± 0.001           \\
SDEdit\cite{SDEdit}                       & 68.51 ± 0.65           & 2.653 ± 0.109                & 0.343 ± 0.001           & 84.58 ± 0.61           & 4.534 ± 0.120                & 0.411 ± 0.001 \\
EGSDE\cite{zhao2022egsde}                 & 59.75 ± 0.62           & \underline{2.028 ± 0.114}                & 0.343 ± 0.001           & 80.17 ± 0.53           & 4.061 ± 0.114                & 0.409 ± 0.001           \\
SDDM\cite{sun2023sddm}                    & 57.38 ± 0.53           & -                & 0.328 ± 0.001           & -                      & -                & -                       \\
CycleDiffusion\cite{wu2023cyclediffusion} & \underline{56.45}        & 2.316                & 0.479         & 69.48                  & \underline{1.253}                & 0.408                   \\
CycleDiff (Ours)                          & \textbf{48.57 ± 0.46}  & \textbf{1.203 ± 0.087}                & \textbf{0.540 ± 0.001}  & \textbf{23.77 ± 0.41}  & \textbf{0.810 ± 0.069}                & \textbf{0.588 ± 0.001}  \\ \hline
\multicolumn{1}{c}{}                      & \multicolumn{3}{c}{Male $\to$ Female}                               & \multicolumn{3}{c}{Female $\to$ Male}                               \\ \hline
CycleGAN\cite{zhu2017cyclegan}            & 67.78                  & -                & -                       & 77.74                  & -                & -                       \\
CUT\cite{park2020cut}                     & 45.03                  & -                & -                       & 47.66                  & -                & -                       \\
Santa\cite{xie2023unpaired}                           & 48.47           & 3.768                & 0.570           & 51.26 & 3.426                & 0.592           \\ \hline
UNSB\cite{UNSB}                           & \textbf{37.87}           & \underline{4.137}                & \underline{0.630}           & 49.43 & 4.436                & \underline{0.628}           \\
SynDiff\cite{ozbey2023unsupervised}   & 108.10                & 10.88              & 0.568                 & 117.15               & 11.31                  & 0.527      \\
ILVR\cite{ILVR}                           & 46.12 ± 0.33           & 4.737 ± 0.142                & 0.510 ± 0.001           & 68.66 ± 0.37           & 6.397 ± 0.131               & 0.443 ± 0.001           \\
SDEdit\cite{SDEdit}                       & 49.43 ± 0.47           & 4.621 ± 0.159                & 0.572 ± 0.000 & 77.56 ± 0.47           & 8.059 ± 0.126                & 0.497 ± 0.001           \\
EGSDE\cite{zhao2022egsde}                 & 45.12 ± 0.24           & 4.596 ± 0.096                & 0.512 ± 0.001           & 72.63 ± 0.19           & 7.094 ± 0.092                & 0.487 ± 0.001           \\
SDDM\cite{sun2023sddm}                    & 44.37 ± 0.23 & -                & 0.526 ± 0.001           & -                      & -                & -                       \\
CycleDiffusion\cite{wu2023cyclediffusion} & 45.04                  & 4.511                & 0.563                   & \underline{46.23}        & \underline{3.607}                & 0.553         \\
CycleDiff (Ours)                          & \underline{43.48 ± 0.18}  & \textbf{3.642 ± 0.052}                & \textbf{0.640 ± 0.001}  & \textbf{44.29 ± 0.24}  & \textbf{2.614 ± 0.047}                & \textbf{0.652 ± 0.001}  \\ \hline
\end{tabular}
}
\vspace{-0.45cm}
\end{center}
\end{table*}

%% file: tables/tab_add.tex
\begin{table*}[ht]
\begin{center}
\caption{
{Quantitative results on additional datasets. The best results are shown in \textbf{bold}, the second-best results are \underline{underlined}, and the third-best results are \underline{\underline{{double-underlined}}}.
}}\label{tab_additional_exp}
\normalsize
\setlength{\tabcolsep}{10pt}
\renewcommand{\arraystretch}{1.15}
\resizebox{\textwidth}{!}{
{
    \begin{tabular}{lcccccccccc}
    \hline
    \multirow{2}{*}{Model} & \multicolumn{2}{c}{\textbf{Old$\to$Young}} & \multicolumn{2}{c}{\textbf{Summer$\to$Winter}} & \multicolumn{2}{c}{\textbf{Label$\to$Cityscape}} & \multicolumn{2}{c}{\textbf{Map$\to$Satellite}} & \multicolumn{2}{c}{\textbf{Horse$\to$Zebra}} \\ \cline{2-11} 
                           & FID$\downarrow$                 & KID$\downarrow$                & FID$\downarrow$                  & KID$\downarrow$                 & FID$\downarrow$                   & KID$\downarrow$                  & FID$\downarrow$                  & KID$\downarrow$                 & FID$\downarrow$                & KID$\downarrow$               \\ \hline
    NOT\cite{NOT}                    & -               & -              & 185.5                & 8.732               & 221.3                 & 19.76                & 224.9                & 16.59               & 104.3                  & 5.012                 \\
    CycleGAN\cite{zhu2017cyclegan}               & 43.5                & -              & 84.9                 & 1.022               & 76.3                  & 3.532                & 54.6                 & 3.430               & 77.2                  & 1.957                 \\
    MUNIT\cite{huang2018munit}                  & -               & -              & 115.4                & 4.901               & 91.4                  & 6.401                & 181.7                & 12.03               & 133.8                  & 3.790                 \\
    Distance\cite{benaim2017distance}               & -                & -              & 97.2                 & 2.843               & 81.8                  & 4.410                & 98.1                 & 5.789               & 72.0                  & 1.856                 \\
    GcGAN\cite{fu2019gcgan}                  & -                & -              & 97.5                 & 2.755               & 105.2                 & 6.824                & 79.4                 & 5.153               & 86.7                  & 2.051                 \\
    CUT\cite{park2020cut}                    & 44.2                & -              & 84.3                 & 1.207               & 56.4                  & 1.611                & 56.1                 & 3.301               & 45.5                  & \textbf{0.541}                 \\
    SDEdit\cite{SDEdit}                 & -                & -              & 118.6                & 3.218               & -                     & -                    & -                    & -                   & 97.3                  & 4.082                 \\
    P2P\cite{P2P}                    & -                & -              & 99.7                 & 2.626               & -                     & -                    & -                    & -                   & 60.9                  & 1.093                 \\
    Santa\cite{xie2023unpaired}                    & 41.9                & -                  & -                    & -                   & 46.1                  & -                    & -                    & -                   & \underline{36.2}               & -                 \\
    UNSB\cite{UNSB}                   & -                & -              & 73.9                 & \textbf{0.421}               & 53.2                  & 1.191                & \textbf{47.6}                 & \textbf{2.013}               & \textbf{35.7}                  & 0.587                 \\
    CycleDiff (Ours)                   & \textbf{40.7}                   & \textbf{1.06}                  & \textbf{72.7}                    & \underline{1.015}                   & \textbf{45.1}                     & \textbf{0.866}                    & \underline{53.2}                    & \underline{3.129}                   & \underline{\underline{43.8}}                  & \underline{0.575}                 \\ \hline
    \end{tabular}
}
}
\vspace{-6pt}
\end{center}
\end{table*}

%% file: sections/exp.tex
\section{Experiment}
\subsection{Experimental setup}
\textbf{Datasets.} We evaluate our method on the following datasets with the resolution of 256 × 256, except for RGB$\leftrightarrow$Depth task with size 375 $\times$ 375: \\
(1) AFHQ \cite{afhq} is a high-quality animal faces dataset, comprising three domains: cat, dog, and wild. Each category has 500 testing images. We conduct experiments on four tasks: Cat$\to$Dog, Dog$\to$Cat, Wild$\to$Dog and Dog$\to$Wild.
(2) CelebA-HQ \cite{celeba} is a high-quality human face dataset and contains two categories, male and female. For each category, there are 1,000 images for validation. We conduct experiments on the dataset with Male$\to$Female and Female$\to$Male. 
(3) Edge2shoes \cite{isola2017image} contains 50,025 shoe images and corresponding edge images. we use 49,825 images for training and 200 images for testing on Edge$\to$RGB and RGB$\to$Edge tasks. 
(4) CelebAMask-HQ \cite{lee2020maskgan} is a high-quality human face segmentation dataset, consisting of 24,183 images for training and 2,824 images for testing. Each image has a corresponding segmentation mask of human facial attributes with 19 classes. 
(5) Virtual KITTI 2 \cite{cabon2020virtual} is a more photo-realistic version of virtual KITTI \cite{gaidon2016virtual} dataset, which contains multiple sets of images such as RGB, depth, class segmentation and so on. The dataset is partitioned into training and testing subsets at a ratio of 1:0.14, with 1,827 images for training and 299 images for testing.
(6) Additional unpaired image-to-image tranlation benchmarks, including: Old$\to$Young \cite{UNSB}, Summer$\to$Winter \cite{zhu2017cyclegan}, Label$\to$Cityscape \cite{cordts2016cityscapes}, Map$\to$Satellite \cite{zhu2017cyclegan} and Horse$\to$Zebra \cite{zhu2017cyclegan}. 

In particular, we conduct experiments on AFHQ and CelebA-HQ datasets for RGB$\leftrightarrow$RGB tasks. Furthermore, the Edge2shoes, CelebAMask-HQ and Virtual KITTI datasets are utilized for the cross-modality tasks of RGB$\leftrightarrow$Edge, RGB$\leftrightarrow$Semantics and RGB$\leftrightarrow$Depth respectively. 

\textbf{Evaluation Metrics.} On RGB$\leftrightarrow$RGB, we evaluate our translated images under two metrics.
To evaluate the realism between translated images and target domain images, we report Frechet Inception Score (FID) \cite{fid} and Kernel Inception Distance (KID) \cite{KID}. To quantify the faithfulness between source domain images and translated images, we report the Structural Similarity Index Measure (SSIM) \cite{ssim}. Following prior work \cite{ye2024diffusionedge}, we compute the Optimal Dataset Scale (ODS) and Optimal Image Scale (OIS) to evaluate the F-scores for general edge detection on RGB$\to$Edge. Following \cite{zheng2022general, te2020edge}, we use the mean F1-score and mean intersection over union (mIOU) overall categories excluding background to measure the semantics performance on RGB$\to$Semantics. Following \cite{patil2022p3depth}, we adopt the Root Mean Squared Error (RMSE) metrics to estimate how well the predicted depths match the ground truth on RGB$\to$Depth.

\subsection{Comparisons to State-of-the-art methods}
{We compare CycleDiff with several state-of-the-art image-to-image translation methods: GAN-based (CycleGAN \cite{zhu2017cyclegan}, CUT \cite{park2020cut}, Santa \cite{xie2023unpaired}), SBDM-based (SDEdit \cite{SDEdit}, ILVR \cite{ILVR}, EGSDE \cite{zhao2022egsde}, SDDM \cite{sun2023sddm}, UNSB \cite{UNSB}) and SynDiff \cite{ozbey2023unsupervised}}. We report their performances across six tasks, detailed in Tab.~\ref{tab_all_public} and Fig.~\ref{fig_all_public}.
The results on Cat$\to$Dog, Wild$\to$Dog and Male$\to$Female are reported from \cite{zhao2022egsde}, \cite{sun2023sddm}, \cite{wu2023cyclediffusion} and \cite{han2021dclgan}. The results on Dog$\to$Cat, Dog$\to$Wild, Female$\to$Male are reproduced by ourselves using their official codes. The results on Horse$\to$Zebra, Summer$\to$Winter, Lable$\to$Cityscape, Map$\to$Satellite and Old$\to$Young are reported from \cite{UNSB} and \cite{xie2023unpaired}. In particular, both ILVR and SDEdit employ 1000 denoising steps, CycleDiffusion utilizes 800 denoising steps, and EGSDE adopts 200 denoising steps. Notably, our method follows SDDM and utilizes only 100 denoising steps.

\textbf{Results on Cat$\leftrightarrow$Dog.} 
As shown in Tab.~\ref{tab_all_public}, CycleDiff outperforms all GAN-based methods by a large margin in the FID metric, which demonstrates the advantage of considering the image translation as a multi-step mapping. 
Especially, CycleDiff exceeds the second best method by 2.32 and 19.61 on Cat$\to$Dog and Dog$\to$Cat respectively.

Note that the SSIM metric denotes the similarity between the translated image and the input image, and a higher score can not mean a better translation performance. For example, CUT gets the best SSIM but the translated image has little changes compared to the input image as shown in Fig.~\ref{fig_all_public}.
Compared to diffusion-based methods, CycleDiff benefits from the joint learning of translation and diffusion, achieving the best FID score and a comparable SSIM to the state of the art. 
We also visualize the qualitative comparisons in Fig.~\ref{fig_all_public}. Our method exhibits more reasonable and structure-consistent outcomes, indicating the superiority of joint learning.

\textbf{Results on Wild$\leftrightarrow$Dog.} 
As reported in the middle of Tab.~\ref{tab_all_public}, our method obtains the best results on the FID metric and very competitive performances on the SSIM metric. 
Compared to GAN-based methods, we outperform CycleGAN and CUT by a large margin (35.38 and 44.37) on the FID metric for Wild$\to$Dog. 
Moreover, our method obtains the best metrics for both Wild$\to$Dog and Dog$\to$Wild, i.e., CycleDiff exceeds the second best method by 7.88 and 19.67 for Wild$\to$Dog and Dog$\to$Wild on FID metric.
Besides, as shown in Fig.~\ref{fig_all_public}, CycleDiff can generate images possessing both structure consistency and high fidelity.

\begin{figure}[t]
    \centering
    \includegraphics[width=0.49\textwidth]{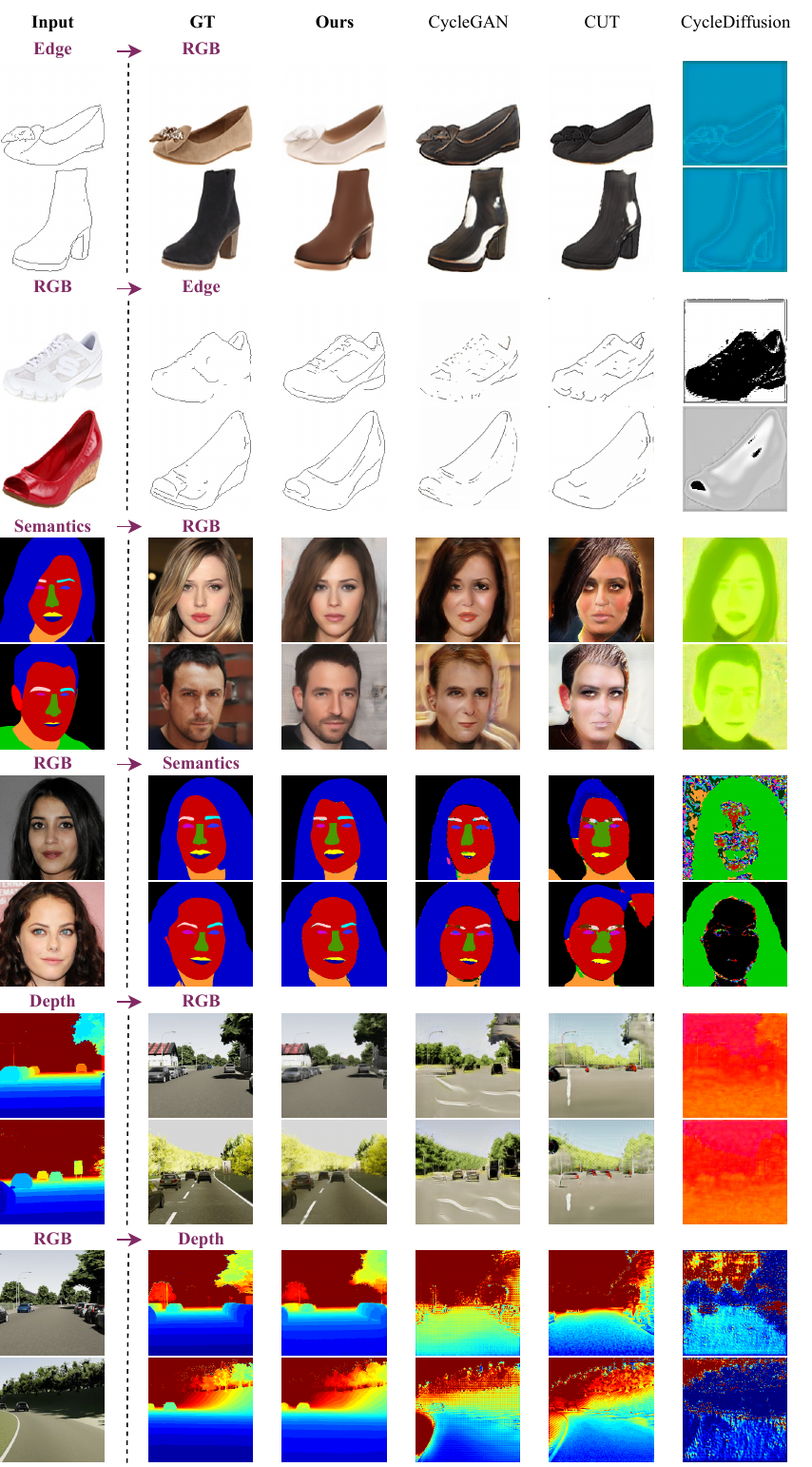}
    \caption{Qualitative comparison on RGB$\leftrightarrow$Edge, RGB$\leftrightarrow$Semantics, RGB$\leftrightarrow$Depth with state-of-the-art methods. CycleDiff is capable of translating images between various modalities. }
    \label{fig_modality}
    \vspace{-5pt}
\end{figure}

\textbf{Results on Male$\leftrightarrow$Female.} 
The bottom of Tab.~\ref{tab_all_public} reports the performance comparisons between CycleDiff and state-of-the-art methods.
CycleDiff achieves the best results on all metrics and exceeds the second best method by 1.94 for Female$\to$Male on the FID metric.
Though the improvement in FID is insignificant compared to other diffusion-based methods, the visual results depicted at the top of Fig.~\ref{fig_all_public} show that our method can maintain better structural consistency. This indicates that the proposed joint learning can improve the generative capability via the improved optimality, resulting in a higher structural similarity.

\begin{figure}[t]
    \centering
    \includegraphics[width=0.47\textwidth]{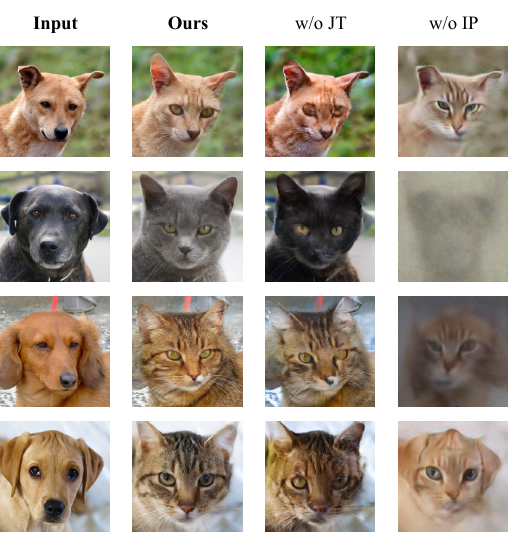}
    \caption{Ablation study on joint training, multi-step mapping and image component. `w/o JT' and `w/o IP' mean `without joint training' and `without image component' respectively.}
    \label{fig_ablation_image_component}
    \vspace{-5pt}
\end{figure}

\textbf{Results on additional benchmarks.} As shown in Tab.~\ref{tab_additional_exp}, our method achieves the best FID scores on Old$\to$Young, Summer$\to$Winter and Lable$\to$Cityscapes tasks and obtains competitive performance to state-of-the-art methods on Map$\to$Satellite and Horse$\to$Zebra tasks. As depicted in Fig.~\ref{fig_additional_dataset}, CycleDiff exhibits exceptional generative power for unpaired image-to-image translation, not only on time-varying datasets such as Old$\to$Young and Summer$\to$Winter, but also on challenging artificial domain data, including Map$\to$Satellite and Label$\to$Cityscape.

\input{tables/tab_cross_modality}

    \begin{table}[t]
    \begin{center}
    \caption{Quantitative comparison results of ablation studies on architectures of the translation network and denoising steps. We conduct experiments on Semantics$\to$RGB. Note that the KID metric is multiplied by 100.
    }\label{tab_ab_archi}
    \setlength{\tabcolsep}{15pt}
    \renewcommand{\arraystretch}{1.15}
    {
    \begin{tabular}{l|cc}
    \hline
    \textbf{Generator Arichitecture}      & FID $\downarrow$ & KID $\downarrow$ \\ \hline
    UNet-arch     & 272.69   & 35.097 \\
    ResNet6-arch  & 80.87   & 9.124 \\
    ResNet9-arch  & 77.42   & 8.073 \\
    ResNet12-arch & \textbf{69.29}   & \textbf{5.975} \\ \hline \hline
    \textbf{Step (with ResNet12-arch)} & FID $\downarrow$ & KID $\downarrow$ \\ \hline
    Step 50                       & 97.39   & 10.079   \\
    Step 100                      & 69.95   & 6.294   \\
    Step 200                      & 69.29   & 5.975   \\
    Step 500                      & 66.83   & 5.694   \\
    Step 800                       & \textbf{65.43}   & \textbf{5.511}   \\ \hline
    \end{tabular}
    }
    \vspace{-5pt}
    \end{center}
    \end{table}

\subsection{Cross-modality Translation}
To demonstrate the effectiveness of our method, we conduct a series of experiments on various modality translation tasks, including RGB$\leftrightarrow$Edge, RGB$\leftrightarrow$Semantics, and RGB$\leftrightarrow$Depth. Specifically, we compare CycleDiff with GAN-based methods including CycleGAN and CUT. We also reproduce the diffusion-based method CycleDiffusion for comparisons of cross-modality tasks and employ 200 denoising steps for generation.

\textbf{Results on RGB$\leftrightarrow$Edge}
Our method achieves more realistic visual results compared to CUT and CycleGAN on Edge$\to$RGB, generating images more closely to ground truth. For RGB$\to$Edge, our method is even able to capture finer details than ground truth, as exemplified in the fourth row of the results. On the other hand, CycleDiffusion performs poorly on cross-modal tasks, highlighting the inherent difficulty of directly transferring between two different domains without joint training.

    \begin{table}[t]
    \begin{center}
    \caption{Quantitative comparison results of ablation studies on the loss component, joint learning and time-dependent translation network.
    }\label{tab_ab_loss}
    \setlength{\tabcolsep}{15pt}
    \renewcommand{\arraystretch}{1.15}
    \resizebox{1\linewidth}{!}
    {
    \begin{tabular}{lccc}
    \hline
    \textbf{Model} & $\mathcal{L}_{dcl}$ & $\mathcal{L}_{lps}$ & FID $\downarrow$ \\ \hline
    w/o D    & -      & $\checkmark$             & 29.13   \\
    w/o P    & $\checkmark$      & -             & 28.21   \\
    Ours      & $\checkmark$      & $\checkmark$    & \textbf{26.45} \\ \hline \hline
    w/o JT & $\checkmark$      & $\checkmark$    & 52.81   \\
    Ours  & $\checkmark$      & $\checkmark$    & \textbf{26.45}   \\ \hline
    w/o TD & $\checkmark$      & $\checkmark$    & 54.31   \\
    w/ TD \& w/o SA & $\checkmark$      & $\checkmark$    & 29.58 \\
    Ours  & $\checkmark$      & $\checkmark$    & \textbf{26.45}   \\ \hline
    \end{tabular}
    }
    \vspace{-5pt}
    \end{center}
    \end{table}

\textbf{Results on RGB$\leftrightarrow$Semantics}
Compared to CycleGAN, CycleDiff produces high-quality results with an improved FID score of 31.76. Concurrently, CycleDiff can generate more precise semantic labels in challenging datasets that encompass 19 categories on RGB$\to$Semantics, demonstrating that our method could align the different modalities with the help of joint training.

\textbf{Results on RGB$\leftrightarrow$Depth}
As shown in Tab.~\ref{tab_downstream}, our method outperforms the state-of-the-art methods by 83.09 FID score on Depth$\leftrightarrow$RGB and achieves the best RMSE of 0.52 on RGB$\leftrightarrow$Depth. As illustrated at the bottom of Fig.~\ref{fig_modality}, the results generated by our method are even close to the ground truth on RGB$\to$Depth and Depth$\to$RGB.

\subsection{Ablation Studies}\label{ablation}

\begin{figure}[t]
    \centering
    \includegraphics[width=0.49\textwidth]{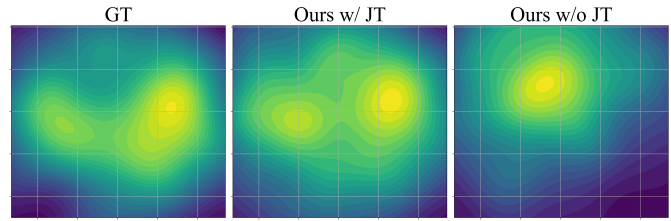}
    \caption{Comparison of distribution under different variants: target domain data distribution, data distribution generated by our method w/ JT and our method w/o JT. We generate 500 images of the target domain on Dog$\to$Cat and adopt Inception-V3~\cite{szegedy2016rethinking} for feature extraction. We then apply Principal Component Analysis (PCA)~\cite{hotelling1933analysis} to reduce the feature dimensionality to two, and use Kernel Density Estimation (KDE)~\cite{epanechnikov1969non} to visualize the probability density of the data distribution.}
    \label{fig_jt}
\end{figure}

\textbf{Effect of the joint training.} To verify the advantages of joint training, we implement a separate training scheme for diffusion models and the cycle translator. In practice, we first train the diffusion models and then utilize the generated images by diffusion models to train the translation process. The setting without joint training is denoted as `w/o JT'.
As reported in Tab.~\ref{tab_ab_loss}, the joint training framework achieves an improvement of FID score by 26.36, enhancing the generated image quality and realistic greatly. As depicted in Fig.~\ref{fig_ablation_image_component}, the joint training manner could synthesize reasonable images with much higher fidelity and structural consistency, demonstrating its effectiveness. 
To further validate the effectiveness of our method, we conduct comparative analysis of feature distributions across different variants using Principal Component Analysis (PCA) and Kernel Density Estimation (KDE). As depicted in Fig.~\ref{fig_jt}, the distribution generated by our method with joint training (w/ JT) closely aligns with the target domain (GT). 
As a comparison, the variant without joint training (w/o JT) shows a noticeable divergence from the target distribution, indicating that the proposed joint training strategy enhances global optimality.

\begin{figure}[t]
    \centering
    \includegraphics[width=0.48\textwidth]{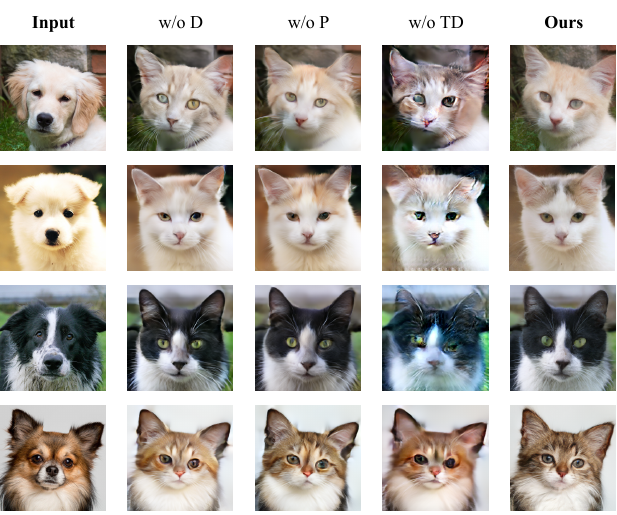}
    \caption{Comparison between different loss components and time-dependent translation network. `w/o D', `w/o P' and `w/o TD' denote `without $\mathcal{L}_{dcl}$', `without $\mathcal{L}_{lps}$' and `w/o time-dependent translation network' respectively.}
    \label{fig_ablation_loss_T}
\end{figure}

\textbf{Effect of the image component.}
In the setting of joint learning, we additionally predict the image component via the denoising network, thus we can combine the translation and diffusion processes.
To show the effect of the image component, we remove the image component, instead, we utilize the estimated noise to calculate a clean image via the forward equation of the diffusion model. In this way, we can use the calculated clean image to conduct the translation process.
As shown in the rightmost column of Fig.~\ref{fig_ablation_image_component}, it can not synthesize the correct images without the image component, demonstrating the importance of extracting image components for effective domain translation.

\textbf{Effect of time-dependent translation network.}
To evaluate the effectiveness of the time-dependent translation network, we modify the time-dependent translation network, removing the time input and just using the encoder-decoder architecture to process the image component. We denote the settings without the time-dependent translation network as `w/o TD'. In addition, we also remove the self-attention layer in the time-attention block to verify its effectiveness and denote this setting as `w/ TD \& w/o SA'.
As reported in Tab.~\ref{tab_ab_loss}, the time-dependent translation network significantly improves the FID by 27.86, which demonstrates its superiority. The self-attention layer could further enhance the fused feature, improving the FID metric by 3.13. We also depict the visual results in Fig.~\ref{fig_ablation_loss_T}, revealing that the time-dependent translation network can yield more realistic and high-quality visual outcomes.

\textbf{Variants of the translation network.}
We find that the network layer of the encoder-decoder in the translation network has a significant impact on the generation performance.
In practice, we ablate the layers of the Resnet blocks and replace the Resnet block with UNet block \cite{ronneberger2015u}. We name these variants UNet-arch, ResNet6-arch (6 layers), ResNet9-arch (9 layers), and ResNet12-arch (12 layers).
As reported in Tab.~\ref{tab_ab_archi}, the ResNet12-arch achieves the best FID score of 69.29. The UNet-arch performs poorly, which indicates the operator of the skip concatenation mechanism may not be suitable for image translation tasks.

\textbf{Effect of the denoising steps.} As shown in Tab.~\ref{tab_ab_archi}, we ablate five different steps ranging from 50 to 800 to evaluate the effect of denoising steps. Furthermore, it surpasses the performance of CycleGAN on the cross-modal task in just 50 steps. Adding the denoising steps requires more time and computation, so we adopt 200 denoising steps as the standard settings for cross-modality image translation experiments.

\textbf{Effect of perceptual and DCL loss.} To explore the effect of the perceptual loss and DCL loss, we analyze CycleDiff by comparing three different settings on Cat$\to$Dog: 1) without DCL loss $\mathcal{L}_{dcl}$ (w/o D), 2) without perceptual loss $\mathcal{L}_{lps}$ (w/o P), 3) with total loss (Ours). 
Tab.~\ref{tab_ab_loss} shows that $\mathcal{L}_{dcl}$ could play an important role in the translation process to ensure the data distribution of translated image components to be close to the target domain and improve the FID by 2.68. The $\mathcal{L}_{lps}$ increases the capability of the generator and produces an improvement of FID by 1.76. 
More specifically, the perceptual loss could preserve the texture information of the input image as shown in the second row of Fig.~\ref{fig_ablation_loss_T} and the DCL loss encourages the source image to be more similar to target domain images as depicted in the last row of Fig.~\ref{fig_ablation_loss_T}.

\input{tables/tab_compute_cost}

\textbf{Comparison of computational cost}.
To evaluate the efficiency of the proposed method, we report the inference time and CUDA memory usage during generation with a batch size of 1, as presented in Tab.~\ref{tab_compute_cost}. Our method demonstrates significantly faster sampling speeds compared to existing state-of-the-art approaches, achieving nearly 7$\times$ acceleration over SDEdit and better performance among diffusion-based methods.

%% file: tables/tab_cross_modality.tex
\begin{table*}[ht]
\begin{center}
\caption{Quantitative comparison on cross-modality tasks with state-of-the-art methods. The best results using diffusion models are in \textbf{bold}.
}\label{tab_downstream}
\normalsize
\setlength{\tabcolsep}{4pt}
\renewcommand{\arraystretch}{0.95}
\resizebox{\textwidth}{!}{
{
\begin{tabular}{l|cccc|cccc|ccc}
\hline
\multirow{2}{*}{} & \multicolumn{2}{c}{Edge$\to$RGB}       & \multicolumn{2}{c|}{RGB$\to$Edge} & \multicolumn{2}{c}{Semantics$\to$RGB} & \multicolumn{2}{c|}{RGB$\to$Semantics} & \multicolumn{2}{c}{Depth$\to$RGB} & RGB$\to$Depth \\
                  & FID $\downarrow$ & KID $\downarrow$ & ODS $\uparrow$ & OIS $\uparrow$ & FID $\downarrow$   & KID $\downarrow$  & mean F1-score (\%) $\uparrow$ & mIOU (\%) $\uparrow$   & FID $\downarrow$      & KID $\downarrow$ & RMSE $\downarrow$     \\ \hline
CycleGAN          & 119.90   & 7.025        & 0.580           & 0.589          & 101.05            & 8.947 & 21.78            & 14.65         & 226.34  & 21.897 & 1.35      \\
CUT               & 99.57  & 5.010          & 0.331          & 0.331       & 107.16 & 9.338           & 17.55            & 11.73         & 233.71  & 24.054 & 1.51       \\
CycleDiffusion       & 381.35 & 48.060  & 0.141 &  0.176    & 260.96  & 41.222          & 3.33            & 1.78         & 464.72  & 61.283 & 2.43       \\
CycleDiff (Ours)               & \textbf{74.93}  & \textbf{2.377}  & \textbf{0.766} &  \textbf{0.769}    & \textbf{69.29}   & \textbf{5.975}         & \textbf{42.95}            & \textbf{30.63}         & \textbf{143.25} & \textbf{12.368} & \textbf{0.52}       \\ \hline
\end{tabular}
}
}
\vspace{-5pt}
\end{center}
\end{table*}

%% file: tables/tab_compute_cost.tex
\begin{table}[t]
\begin{center}
\caption{
{Comparison with the state-of-the-art methods in terms of inference speed and CUDA memory usage.}}\label{tab_compute_cost}
\vspace{0.1cm}
\normalsize
\setlength{\tabcolsep}{5pt}
\renewcommand{\arraystretch}{0.95}
{
\begin{tabular}{l|ccl}
\hline
Method         & sec/iter $\downarrow$      & CUDA Memory (GB) $\downarrow$ \\ \hline
CUT           & 0.24 & 2.91 \\
ILVR           &  60   &  1.84     \\
SDEdit         & 33   &   2.20    \\
EGSDE          & 85 &  3.64  \\
CycleDiffusion & 46  & 4.23 \\
CycleDiff (Ours) & 4.9  & 9.31 \\ \hline
\end{tabular}
}
\end{center}
\end{table}

%% file: sections/conclu.tex
\section{Conclusion {and Future Work}}
This paper introduces CycleDiff, a diffusion-based translator for unpaired image-to-image translation tasks. We are the first attempt to learn the diffusion and translation processes jointly. On the one hand, we propose to extract the image components from diffusion models to allow for joint learning. On the other hand, we introduce a time-dependent translation network to learn the translation process efficiently. Our method can be easily applied to cross-modality image translation tasks, including RGB$\to$Edge, RGB$\to$Semantics and RGB$\to$Depth. We conduct experiments on five different public datasets, showcasing the great superiority.

{In the future, one direction that could be investigated is to extend our method to more domains, such as RGB$\leftrightarrow$Norm, RGB$\leftrightarrow$Text, and medical image translation such as MRI$\leftrightarrow$CT. An additional promising research direction involves leveraging our cycle translation framework to bridge the sim-to-real domain gap in robotic learning applications.}

%% file: sections/ack.tex
\section{Acknowledgment}
We thank the anonymous reviewers for their valuable comments. This work is supported in part by the NSFC (62522219), the Major Program of Xiangjiang Laboratory (23XJ01009), NSFC (62325211, 62132021, 62372457, 62572477).

%% file: main.bbl
\begin{thebibliography}{10}
\providecommand{\url}[1]{#1}
\csname url@samestyle\endcsname
\providecommand{\newblock}{\relax}
\providecommand{\bibinfo}[2]{#2}
\providecommand{\BIBentrySTDinterwordspacing}{\spaceskip=0pt\relax}
\providecommand{\BIBentryALTinterwordstretchfactor}{4}
\providecommand{\BIBentryALTinterwordspacing}{\spaceskip=\fontdimen2\font plus
\BIBentryALTinterwordstretchfactor\fontdimen3\font minus \fontdimen4\font\relax}
\providecommand{\BIBforeignlanguage}[2]{{%
\expandafter\ifx\csname l@#1\endcsname\relax
\typeout{** WARNING: IEEEtran.bst: No hyphenation pattern has been}%
\typeout{** loaded for the language `#1'. Using the pattern for}%
\typeout{** the default language instead.}%
\else
\language=\csname l@#1\endcsname
\fi
#2}}
\providecommand{\BIBdecl}{\relax}
\BIBdecl

\bibitem{richardson2021encoding}
E.~Richardson, Y.~Alaluf, O.~Patashnik, Y.~Nitzan, Y.~Azar, S.~Shapiro, and D.~Cohen-Or, ``Encoding in style: a stylegan encoder for image-to-image translation,'' in \emph{Proceedings of the IEEE/CVF conference on computer vision and pattern recognition}, 2021, pp. 2287--2296.

\bibitem{wang2020multi}
Y.~Wang, Z.~Zhang, W.~Hao, and C.~Song, ``Multi-domain image-to-image translation via a unified circular framework,'' \emph{IEEE Transactions on Image Processing}, pp. 670--684, 2020.

\bibitem{ye2024diffusionedge}
Y.~Ye, K.~Xu, Y.~Huang, R.~Yi, and Z.~Cai, ``Diffusionedge: Diffusion probabilistic model for crisp edge detection,'' in \emph{Proceedings of the AAAI Conference on Artificial Intelligence}, 2024, pp. 6675--6683.

\bibitem{liu2020adaptive}
Y.~Liu, Z.~Xie, and H.~Liu, ``An adaptive and robust edge detection method based on edge proportion statistics,'' \emph{IEEE Transactions on Image Processing}, pp. 5206--5215, 2020.

\bibitem{zhou2021gmnet}
W.~Zhou, J.~Liu, J.~Lei, L.~Yu, and J.-N. Hwang, ``Gmnet: Graded-feature multilabel-learning network for rgb-thermal urban scene semantic segmentation,'' \emph{IEEE Transactions on Image Processing}, pp. 7790--7802, 2021.

\bibitem{wu2023conditional}
D.~Wu, Z.~Guo, A.~Li, C.~Yu, C.~Gao, and N.~Sang, ``Conditional boundary loss for semantic segmentation,'' \emph{IEEE Transactions on Image Processing}, 2023.

\bibitem{ye2021unsupervised}
X.~Ye, X.~Fan, M.~Zhang, R.~Xu, and W.~Zhong, ``Unsupervised monocular depth estimation via recursive stereo distillation,'' \emph{IEEE Transactions on Image Processing}, pp. 4492--4504, 2021.

\bibitem{xu2021multi}
X.~Xu, Z.~Chen, and F.~Yin, ``Multi-scale spatial attention-guided monocular depth estimation with semantic enhancement,'' \emph{IEEE Transactions on Image Processing}, pp. 8811--8822, 2021.

\bibitem{zhu2017cyclegan}
J.-Y. Zhu, T.~Park, P.~Isola, and A.~A. Efros, ``Unpaired image-to-image translation using cycle-consistent adversarial networks,'' in \emph{Proceedings of the IEEE international conference on computer vision}, 2017, pp. 2223--2232.

\bibitem{yi2017dualgan}
Z.~Yi, H.~Zhang, P.~Tan, and M.~Gong, ``Dualgan: Unsupervised dual learning for image-to-image translation,'' in \emph{Proceedings of the IEEE international conference on computer vision}, 2017, pp. 2849--2857.

\bibitem{kim2017discogan}
T.~Kim, M.~Cha, H.~Kim, J.~K. Lee, and J.~Kim, ``Learning to discover cross-domain relations with generative adversarial networks,'' in \emph{International conference on machine learning}, 2017, pp. 1857--1865.

\bibitem{kurz2016kullback}
G.~Kurz, F.~Pfaff, and U.~D. Hanebeck, ``Kullback-leibler divergence and moment matching for hyperspherical probability distributions,'' in \emph{2016 19th International Conference on Information Fusion (FUSION)}, 2016, pp. 2087--2094.

\bibitem{zhao2022egsde}
M.~Zhao, F.~Bao, C.~Li, and J.~Zhu, ``Egsde: Unpaired image-to-image translation via energy-guided stochastic differential equations,'' \emph{Advances in Neural Information Processing Systems}, pp. 3609--3623, 2022.

\bibitem{wu2023cyclediffusion}
C.~H. Wu and F.~De~la Torre, ``A latent space of stochastic diffusion models for zero-shot image editing and guidance,'' in \emph{Proceedings of the IEEE/CVF International Conference on Computer Vision}, 2023, pp. 7378--7387.

\bibitem{sun2023sddm}
S.~Sun, L.~Wei, J.~Xing, J.~Jia, and Q.~Tian, ``Sddm: score-decomposed diffusion models on manifolds for unpaired image-to-image translation,'' in \emph{International Conference on Machine Learning}, 2023, pp. 33\,115--33\,134.

\bibitem{sasaki2021unit}
H.~Sasaki, C.~G. Willcocks, and T.~P. Breckon, ``Unit-ddpm: Unpaired image translation with denoising diffusion probabilistic models,'' \emph{arXiv preprint arXiv:2104.05358}, 2021.

\bibitem{ozbey2023unsupervised}
M.~{\"O}zbey, O.~Dalmaz, S.~U. Dar, H.~A. Bedel, {\c{S}}.~{\"O}zturk, A.~G{\"u}ng{\"o}r, and T.~{\c{C}}ukur, ``Unsupervised medical image translation with adversarial diffusion models,'' \emph{IEEE Transactions on Medical Imaging}, vol.~42, no.~12, pp. 3524--3539, 2023.

\bibitem{isola2017image}
P.~Isola, J.-Y. Zhu, T.~Zhou, and A.~A. Efros, ``Image-to-image translation with conditional adversarial networks,'' in \emph{Proceedings of the IEEE conference on computer vision and pattern recognition}, 2017, pp. 1125--1134.

\bibitem{ugatit}
J.~Kim, M.~Kim, H.~Kang, and K.~Lee, ``{U-GAT-IT:} unsupervised generative attentional networks with adaptive layer-instance normalization for image-to-image translation,'' in \emph{8th International Conference on Learning Representations, {ICLR} 2020, Addis Ababa, Ethiopia, April 26-30, 2020}, 2020.

\bibitem{xie2023unpaired}
S.~Xie, Y.~Xu, M.~Gong, and K.~Zhang, ``Unpaired image-to-image translation with shortest path regularization,'' in \emph{Proceedings of the IEEE/CVF conference on computer vision and pattern recognition}, 2023, pp. 10\,177--10\,187.

\bibitem{dhariwal2021diffusionbeatgan}
P.~Dhariwal and A.~Nichol, ``Diffusion models beat gans on image synthesis,'' \emph{Advances in neural information processing systems}, pp. 8780--8794, 2021.

\bibitem{ozdenizci2023restoring}
O.~{\"O}zdenizci and R.~Legenstein, ``Restoring vision in adverse weather conditions with patch-based denoising diffusion models,'' \emph{IEEE Transactions on Pattern Analysis and Machine Intelligence}, 2023.

\bibitem{luo2023fast}
T.~Luo, Z.~Mo, and S.~J. Pan, ``Fast graph generation via spectral diffusion,'' \emph{IEEE Transactions on Pattern Analysis and Machine Intelligence}, 2023.

\bibitem{lu2022conditional}
Y.-J. Lu, Z.-Q. Wang, S.~Watanabe, A.~Richard, C.~Yu, and Y.~Tsao, ``Conditional diffusion probabilistic model for speech enhancement,'' in \emph{ICASSP 2022-2022 IEEE International Conference on Acoustics, Speech and Signal Processing (ICASSP)}, 2022, pp. 7402--7406.

\bibitem{ho2020ddpm}
J.~Ho, A.~Jain, and P.~Abbeel, ``Denoising diffusion probabilistic models,'' \emph{Advances in neural information processing systems}, pp. 6840--6851, 2020.

\bibitem{nichol2021ddim}
A.~Q. Nichol and P.~Dhariwal, ``Improved denoising diffusion probabilistic models,'' in \emph{International conference on machine learning}, 2021, pp. 8162--8171.

\bibitem{scorebaseddm}
Y.~Song, J.~Sohl{-}Dickstein, D.~P. Kingma, A.~Kumar, S.~Ermon, and B.~Poole, ``Score-based generative modeling through stochastic differential equations,'' in \emph{9th International Conference on Learning Representations, {ICLR} 2021, Virtual Event, Austria, May 3-7, 2021}, 2021.

\bibitem{rombach2022high}
R.~Rombach, A.~Blattmann, D.~Lorenz, P.~Esser, and B.~Ommer, ``High-resolution image synthesis with latent diffusion models,'' in \emph{Proceedings of the IEEE/CVF conference on computer vision and pattern recognition}, 2022, pp. 10\,684--10\,695.

\bibitem{ILVR}
J.~Choi, S.~Kim, Y.~Jeong, Y.~Gwon, and S.~Yoon, ``{ILVR:} conditioning method for denoising diffusion probabilistic models,'' in \emph{2021 {IEEE/CVF} International Conference on Computer Vision, {ICCV} 2021, Montreal, QC, Canada, October 10-17, 2021}, 2021, pp. 14\,347--14\,356.

\bibitem{SDEdit}
C.~Meng, Y.~He, Y.~Song, J.~Song, J.~Wu, J.~Zhu, and S.~Ermon, ``Sdedit: Guided image synthesis and editing with stochastic differential equations,'' in \emph{The Tenth International Conference on Learning Representations, {ICLR} 2022, Virtual Event, April 25-29, 2022}, 2022.

\bibitem{UNSB}
B.~Kim, G.~Kwon, K.~Kim, and J.~C. Ye, ``Unpaired image-to-image translation via neural schr{\"{o}}dinger bridge,'' in \emph{The Twelfth International Conference on Learning Representations, {ICLR} 2024, Vienna, Austria, May 7-11, 2024}.\hskip 1em plus 0.5em minus 0.4em\relax OpenReview.net, 2024.

\bibitem{huangddm}
Y.~Huang, Z.~Qin, X.~Liu, and K.~Xu, ``Simultaneous image to zero and zero to noise: Diffusion models with analytical image attenuation,'' \emph{arXiv preprint arXiv:2306.13720}, 2023.

\bibitem{he2016deep}
K.~He, X.~Zhang, S.~Ren, and J.~Sun, ``Deep residual learning for image recognition,'' in \emph{Proceedings of the IEEE conference on computer vision and pattern recognition}, 2016, pp. 770--778.

\bibitem{perez2018film}
E.~Perez, F.~Strub, H.~De~Vries, V.~Dumoulin, and A.~Courville, ``Film: Visual reasoning with a general conditioning layer,'' in \emph{Proceedings of the AAAI conference on artificial intelligence}, vol.~32, no.~1, 2018.

\bibitem{johnson2016perceptual}
J.~Johnson, A.~Alahi, and L.~Fei-Fei, ``Perceptual losses for real-time style transfer and super-resolution,'' in \emph{Computer Vision--ECCV 2016: 14th European Conference, Amsterdam, The Netherlands, October 11-14, 2016, Proceedings, Part II 14}, 2016, pp. 694--711.

\bibitem{vgg16}
K.~Simonyan and A.~Zisserman, ``Very deep convolutional networks for large-scale image recognition,'' in \emph{3rd International Conference on Learning Representations, {ICLR} 2015, San Diego, CA, USA, May 7-9, 2015, Conference Track Proceedings}, Y.~Bengio and Y.~LeCun, Eds., 2015.

\bibitem{huang2018munit}
X.~Huang, M.-Y. Liu, S.~Belongie, and J.~Kautz, ``Multimodal unsupervised image-to-image translation,'' in \emph{Proceedings of the European conference on computer vision (ECCV)}, 2018, pp. 172--189.

\bibitem{lee2018drit}
H.-Y. Lee, H.-Y. Tseng, J.-B. Huang, M.~Singh, and M.-H. Yang, ``Diverse image-to-image translation via disentangled representations,'' in \emph{Proceedings of the European conference on computer vision (ECCV)}, 2018, pp. 35--51.

\bibitem{benaim2017distance}
S.~Benaim and L.~Wolf, ``One-sided unsupervised domain mapping,'' \emph{Advances in neural information processing systems}, 2017.

\bibitem{fu2019gcgan}
H.~Fu, M.~Gong, C.~Wang, K.~Batmanghelich, K.~Zhang, and D.~Tao, ``Geometry-consistent generative adversarial networks for one-sided unsupervised domain mapping,'' in \emph{Proceedings of the IEEE/CVF conference on computer vision and pattern recognition}, 2019, pp. 2427--2436.

\bibitem{zheng2021lsesim}
C.~Zheng, T.-J. Cham, and J.~Cai, ``The spatially-correlative loss for various image translation tasks,'' in \emph{Proceedings of the IEEE/CVF conference on computer vision and pattern recognition}, 2021, pp. 16\,407--16\,417.

\bibitem{zheng2022ittr}
W.~Zheng, Q.~Li, G.~Zhang, P.~Wan, and Z.~Wang, ``Ittr: Unpaired image-to-image translation with transformers,'' \emph{arXiv preprint arXiv:2203.16015}, 2022.

\bibitem{choi2020starganv2}
Y.~Choi, Y.~Uh, J.~Yoo, and J.-W. Ha, ``Stargan v2: Diverse image synthesis for multiple domains,'' in \emph{Proceedings of the IEEE/CVF conference on computer vision and pattern recognition}, 2020, pp. 8188--8197.

\bibitem{park2020cut}
T.~Park, A.~A. Efros, R.~Zhang, and J.-Y. Zhu, ``Contrastive learning for unpaired image-to-image translation,'' in \emph{Computer Vision--ECCV 2020: 16th European Conference, Glasgow, UK, August 23--28, 2020, Proceedings, Part IX 16}, 2020, pp. 319--345.

\bibitem{NOT}
A.~Korotin, D.~Selikhanovych, and E.~Burnaev, ``Neural optimal transport,'' in \emph{The Eleventh International Conference on Learning Representations, {ICLR} 2023, Kigali, Rwanda, May 1-5, 2023}.\hskip 1em plus 0.5em minus 0.4em\relax OpenReview.net, 2023.

\bibitem{P2P}
A.~Hertz, R.~Mokady, J.~Tenenbaum, K.~Aberman, Y.~Pritch, and D.~Cohen{-}Or, ``Prompt-to-prompt image editing with cross-attention control,'' in \emph{The Eleventh International Conference on Learning Representations, {ICLR} 2023, Kigali, Rwanda, May 1-5, 2023}.\hskip 1em plus 0.5em minus 0.4em\relax OpenReview.net, 2023.

\bibitem{song2020improved}
Y.~Song and S.~Ermon, ``Improved techniques for training score-based generative models,'' \emph{Advances in neural information processing systems}, vol.~33, pp. 12\,438--12\,448, 2020.

\bibitem{mao2017least}
X.~Mao, Q.~Li, H.~Xie, R.~Y. Lau, Z.~Wang, and S.~Paul~Smolley, ``Least squares generative adversarial networks,'' in \emph{Proceedings of the IEEE international conference on computer vision}, 2017, pp. 2794--2802.

\bibitem{afhq}
Y.~Choi, Y.~Uh, J.~Yoo, and J.~Ha, ``Stargan v2: Diverse image synthesis for multiple domains,'' in \emph{2020 {IEEE/CVF} Conference on Computer Vision and Pattern Recognition, {CVPR} 2020, Seattle, WA, USA, June 13-19, 2020}.\hskip 1em plus 0.5em minus 0.4em\relax Computer Vision Foundation / {IEEE}, 2020, pp. 8185--8194.

\bibitem{celeba}
T.~Karras, T.~Aila, S.~Laine, and J.~Lehtinen, ``Progressive growing of gans for improved quality, stability, and variation,'' in \emph{6th International Conference on Learning Representations, {ICLR} 2018, Vancouver, BC, Canada, April 30 - May 3, 2018, Conference Track Proceedings}, 2018.

\bibitem{lee2020maskgan}
C.-H. Lee, Z.~Liu, L.~Wu, and P.~Luo, ``Maskgan: Towards diverse and interactive facial image manipulation,'' in \emph{Proceedings of the IEEE/CVF conference on computer vision and pattern recognition}, 2020, pp. 5549--5558.

\bibitem{cabon2020virtual}
Y.~Cabon, N.~Murray, and M.~Humenberger, ``Virtual kitti 2,'' \emph{arXiv preprint arXiv:2001.10773}, 2020.

\bibitem{gaidon2016virtual}
A.~Gaidon, Q.~Wang, Y.~Cabon, and E.~Vig, ``Virtual worlds as proxy for multi-object tracking analysis,'' in \emph{Proceedings of the IEEE conference on computer vision and pattern recognition}, 2016, pp. 4340--4349.

\bibitem{cordts2016cityscapes}
M.~Cordts, M.~Omran, S.~Ramos, T.~Rehfeld, M.~Enzweiler, R.~Benenson, U.~Franke, S.~Roth, and B.~Schiele, ``The cityscapes dataset for semantic urban scene understanding,'' in \emph{Proceedings of the IEEE conference on computer vision and pattern recognition}, 2016, pp. 3213--3223.

\bibitem{fid}
M.~Heusel, H.~Ramsauer, T.~Unterthiner, B.~Nessler, and S.~Hochreiter, ``Gans trained by a two time-scale update rule converge to a local nash equilibrium,'' in \emph{Advances in Neural Information Processing Systems 30: Annual Conference on Neural Information Processing Systems 2017, December 4-9, 2017, Long Beach, CA, {USA}}, 2017, pp. 6626--6637.

\bibitem{KID}
M.~Binkowski, D.~J. Sutherland, M.~Arbel, and A.~Gretton, ``Demystifying {MMD} gans,'' in \emph{6th International Conference on Learning Representations, {ICLR} 2018, Vancouver, BC, Canada, April 30 - May 3, 2018, Conference Track Proceedings}.\hskip 1em plus 0.5em minus 0.4em\relax OpenReview.net, 2018.

\bibitem{ssim}
Z.~Wang, A.~C. Bovik, H.~R. Sheikh, and E.~P. Simoncelli, ``Image quality assessment: from error visibility to structural similarity,'' \emph{{IEEE} Trans. Image Process.}, vol.~13, no.~4, pp. 600--612, 2004.

\bibitem{zheng2022general}
Y.~Zheng, H.~Yang, T.~Zhang, J.~Bao, D.~Chen, Y.~Huang, L.~Yuan, D.~Chen, M.~Zeng, and F.~Wen, ``General facial representation learning in a visual-linguistic manner,'' in \emph{Proceedings of the IEEE/CVF conference on computer vision and pattern recognition}, 2022, pp. 18\,697--18\,709.

\bibitem{te2020edge}
G.~Te, Y.~Liu, W.~Hu, H.~Shi, and T.~Mei, ``Edge-aware graph representation learning and reasoning for face parsing,'' in \emph{Computer Vision--ECCV 2020: 16th European Conference, Glasgow, UK, August 23--28, 2020, Proceedings, Part XII 16}, 2020, pp. 258--274.

\bibitem{patil2022p3depth}
V.~Patil, C.~Sakaridis, A.~Liniger, and L.~Van~Gool, ``P3depth: Monocular depth estimation with a piecewise planarity prior,'' in \emph{Proceedings of the IEEE/CVF Conference on Computer Vision and Pattern Recognition}, 2022, pp. 1610--1621.

\bibitem{han2021dclgan}
J.~Han, M.~Shoeiby, L.~Petersson, and M.~A. Armin, ``Dual contrastive learning for unsupervised image-to-image translation,'' in \emph{Proceedings of the IEEE/CVF conference on computer vision and pattern recognition}, 2021, pp. 746--755.

\bibitem{szegedy2016rethinking}
C.~Szegedy, V.~Vanhoucke, S.~Ioffe, J.~Shlens, and Z.~Wojna, ``Rethinking the inception architecture for computer vision,'' in \emph{Proceedings of the IEEE conference on computer vision and pattern recognition}, 2016, pp. 2818--2826.

\bibitem{hotelling1933analysis}
H.~Hotelling, ``Analysis of a complex of statistical variables into principal components.'' \emph{Journal of educational psychology}, vol.~24, no.~6, p. 417, 1933.

\bibitem{epanechnikov1969non}
V.~A. Epanechnikov, ``Non-parametric estimation of a multivariate probability density,'' \emph{Theory of Probability \& Its Applications}, vol.~14, no.~1, pp. 153--158, 1969.

\bibitem{ronneberger2015u}
O.~Ronneberger, P.~Fischer, and T.~Brox, ``U-net: Convolutional networks for biomedical image segmentation,'' in \emph{Medical image computing and computer-assisted intervention--MICCAI 2015: 18th international conference, Munich, Germany, October 5-9, 2015, proceedings, part III 18}, 2015, pp. 234--241.

\end{thebibliography}
